\documentclass[a4paper,fleqn]{cas-dc}

\usepackage[numbers]{natbib}
\usepackage{array}
\usepackage[caption=false,font=normalsize,labelfont=sf,textfont=sf]{subfig}
\usepackage{textcomp}
\usepackage{stfloats}
\usepackage{url}
\usepackage{verbatim}
\usepackage[linesnumbered,ruled,vlined,noend]{algorithm2e}
\usepackage{caption}
\usepackage{ulem}
\usepackage{adjustbox}
\usepackage{float}
\usepackage{diagbox}
\usepackage{color}
\usepackage{colortbl}
\usepackage{needspace}
\usepackage{lipsum}
\usepackage{lastpage}
\usepackage[table]{xcolor}
\definecolor{hiscolor}{RGB}{220,224,228}
\definecolor{applegreen}{rgb}{0.55, 0.71, 0.0}
\definecolor{downred}{RGB}{255, 136, 132}
\definecolor{rolecolor_1}{RGB}{255,247,172}
\definecolor{rolecolor_2}{RGB}{221,241,243}
\definecolor{rolecolor_3}{RGB}{236,244,221}
\definecolor{intercolor_1}{RGB}{255,230,230}
\definecolor{intercolor_2}{RGB}{221,242,247}
\definecolor{bestcolor}{RGB}{142,139,254}
\usepackage{multirow}
\usepackage{enumitem}
\usepackage{makecell}
\usepackage{graphicx}
\usepackage{wrapfig}
\usepackage[utf8]{inputenc}
\usepackage{amsmath}

\definecolor{markcolor}{RGB}{220,224,228}
\usepackage[T1]{fontenc}    
\usepackage{hyperref}       
\usepackage{url}            
\usepackage{booktabs}       
\usepackage{amsfonts}       
\usepackage{nicefrac}       
\usepackage{microtype}      
\usepackage{xcolor}         
\usepackage{soul}
\usepackage[T1]{fontenc}    
\usepackage{hyperref}       
\usepackage{amsfonts}       
\usepackage{nicefrac}       
\usepackage{microtype}      

\usepackage{tcolorbox}

\newcounter{mytcolorbox}

\renewcommand{\themytcolorbox}{\arabic{mytcolorbox}}

\def\tsc#1{\csdef{#1}{\textsc{\lowercase{#1}}\xspace}}
\tsc{WGM}
\tsc{QE}

\begin{document}
\let\WriteBookmarks\relax
\def\floatpagepagefraction{1}
\def\textpagefraction{.001}

\AtBeginDocument{\gdef\lastpage{\pageref{LastPage}}}

\shorttitle{}    

\shortauthors{}  

\title [mode = title]{CoCR-RAG: Enhancing Retrieval-Augmented Generation in Web Q\&A via Concept-oriented Context Reconstruction}  

%

\author[1,2]{Kaize Shi}[orcid=0000-0003-3561-3627]

\ead{Kaize.Shi@unisq.edu.au}

\author[2,3]{Xueyao Sun}[orcid=0000-0003-2212-9422]

\ead{Xueyao.Sun@connect.polyu.hk}

\author[4]{Qika Lin}[orcid=0000-0001-5650-0600]

\ead{linqika@nus.edu.sg}

\author[5]{Firoj Alam}[orcid=0000-0001-7172-1997]

\ead{fialam@hbku.edu.qa}

\author[3]{Qing Li}[orcid=0000-0003-3370-471X]

\ead{qing-prof.li@polyu.edu.hk}

\author[1]{Xiaohui Tao}[orcid=0000-0002-0020-077X]

\ead{Xiaohui.Tao@unisq.edu.au}

\author[6]{Guandong Xu}[orcid=0000-0003-4493-6663]
\cormark[1]

\ead{gdxu@eduhk.hk}

\cortext[1]{Corresponding author}

\affiliation[1]{organization={University of Southern Queensland},
            state={QLD},
            country={Australia}}

\affiliation[2]{organization={University of Technology Sydney},
            state={NSW},
            country={Australia}}

\affiliation[3]{organization={The Hong Kong Polytechnic University},
            state={Hong Kong},
            country={China}}

\affiliation[4]{organization={National University of Singapore},
            state={Singapore},
            country={Singapore}}

\affiliation[5]{organization={Qatar Computing Research Institute},
            state={Doha},
            country={Qatar}}

\affiliation[6]{organization={The Education University of Hong Kong},
            state={Hong Kong},
            country={China}}


\begin{abstract}
Retrieval-augmented generation (RAG) has shown promising results in enhancing Q\&A by incorporating information from the web and other external sources. However, the supporting documents retrieved from the heterogeneous web often originate from multiple sources with diverse writing styles, varying formats, and inconsistent granularity. Fusing such multi-source documents into a coherent and knowledge-intensive context remains a significant challenge, as the presence of irrelevant and redundant information can compromise the factual consistency of the inferred answers. This paper proposes the \underline{C}oncept-\underline{o}riented \underline{C}ontext \underline{R}econstruction \underline{RAG} (CoCR-RAG), a framework that addresses the multi-source information fusion problem in RAG through linguistically grounded concept-level integration. Specifically, we introduce a concept distillation algorithm that extracts essential concepts from Abstract Meaning Representation (AMR), a stable semantic representation that structures the meaning of texts as logical graphs. The distilled concepts from multiple retrieved documents are then fused and reconstructed into a unified, information-intensive context by Large Language Models (LLMs), which supplement only the necessary sentence elements to highlight the core knowledge. This semantic-level fusion strategy transforms scattered and noisy multi-source documents into a compact, concept-oriented representation that reduces interference from irrelevant information while preserving essential factual content. Experiments on the PopQA and EntityQuestions datasets demonstrate that CoCR-RAG significantly outperforms existing context-reconstruction methods across these Web Q\&A benchmarks. Furthermore, CoCR-RAG shows robustness across various backbone LLMs, establishing itself as a flexible, plug-and-play component adaptable to different RAG frameworks. This work introduces a paradigm for reliable multi-source information fusion in RAG through stable semantic features, offering a linguistically grounded solution for integrating heterogeneous web-retrieved documents.
\end{abstract}


\maketitle

\section{Introduction}
\label{sec:intro}

Large Language Models (LLMs) have made remarkable advancements in Natural Language Generation (NLG), especially in their capacity to consistently address a wide range of tasks interactively~\cite{he2025survey}. Through extensive pre-training, LLMs accumulate large amounts of knowledge from diverse sources, allowing them to produce coherent and fluent text across many domains~\cite{shi2025llama}. However, LLMs are susceptible to producing hallucinations and inconsistencies, especially when responding to queries involving long-tail or domain-specific knowledge. This limitation stems from their reliance on fixed parametric memory, which may contain outdated, incomplete, or erroneous information. Addressing this challenge is important for enhancing the reliability and trustworthiness of answers generated by LLMs, where the provision of accurate and up-to-date information is essential for maximizing their practical utility and adaptability~\cite{yang2025integrating}.

Retrieval-Augmented Generation (RAG) has emerged as an effective paradigm to overcome the limitations of LLMs by enabling them to leverage up-to-date information from the web and other external knowledge sources. By incorporating non-parametric knowledge from retrieved supporting documents, RAG expands the knowledge boundaries of LLMs and mitigates the impact of potentially flawed or outdated internal memory. However, the integration of external information introduces new challenges, especially when dealing with noisy or irrelevant content frequently found in heterogeneous web-retrieved documents~\cite{liu2025retrieval}. The abundance of redundant or misleading data can lead to hallucinations, where the model generates answers misaligned with query intent, thus undermining the reliability and accuracy of RAG systems.

A central challenge underlying these issues can be characterized as a multi-source information fusion problem. In a typical Web Q\&A scenario, a retriever collects multiple supporting documents from different web sources, each with its own writing style, level of detail, and information granularity. These documents may contain overlapping facts expressed in different surface forms, contradictory claims, or large amounts of irrelevant content surrounding only a few useful concepts. Effectively fusing the essential knowledge from such heterogeneous multi-source documents into a unified, information-intensive representation is therefore critical for reliable downstream inference~\cite{mao2024survey}. Unlike traditional information fusion settings that operate on simply combining text or feature vectors, RAG-based Web Q\&A requires a semantic-level fusion strategy that can handle unstructured natural language from diverse web sources while preserving the core factual content and filtering out noise.

Linguistic structures, including semantic and syntactic features, have shown substantial promise in enhancing the interpretability, controllability, and diversity of NLG systems~\cite{li-etal-2023-explicit}. By explicitly leveraging these linguistic features, the input can be structured into compact and informative representations, improving overall system performance~\cite{shannon1948mathematical}. Compared to parameterized augmentation methods, linguistically based approaches offer robust and explicit features that enhance interpretability and coherence in text generation~\cite{narayan2021planning}. These methods, grounded in fundamental linguistic elements, are less vulnerable to noise from non-standard expressions, allowing them to more effectively capture key concepts from underlying structures while mitigating biases often present in web-based information~\cite{liu2020incomplete}. In particular, within RAG frameworks, reorganizing core concepts from unstructured web data can significantly reduce the inference burden. This process parallels human cognition, where the reorganization of essential concepts aids in understanding complex documents~\cite{shi2025concept}. Consequently, linguistic structures hold substantial potential for serving as the basis of a semantic-level fusion mechanism that integrates knowledge from heterogeneous sources into coherent contexts for reasoning.

\begin{figure}
\centering
\includegraphics[width=1\linewidth]{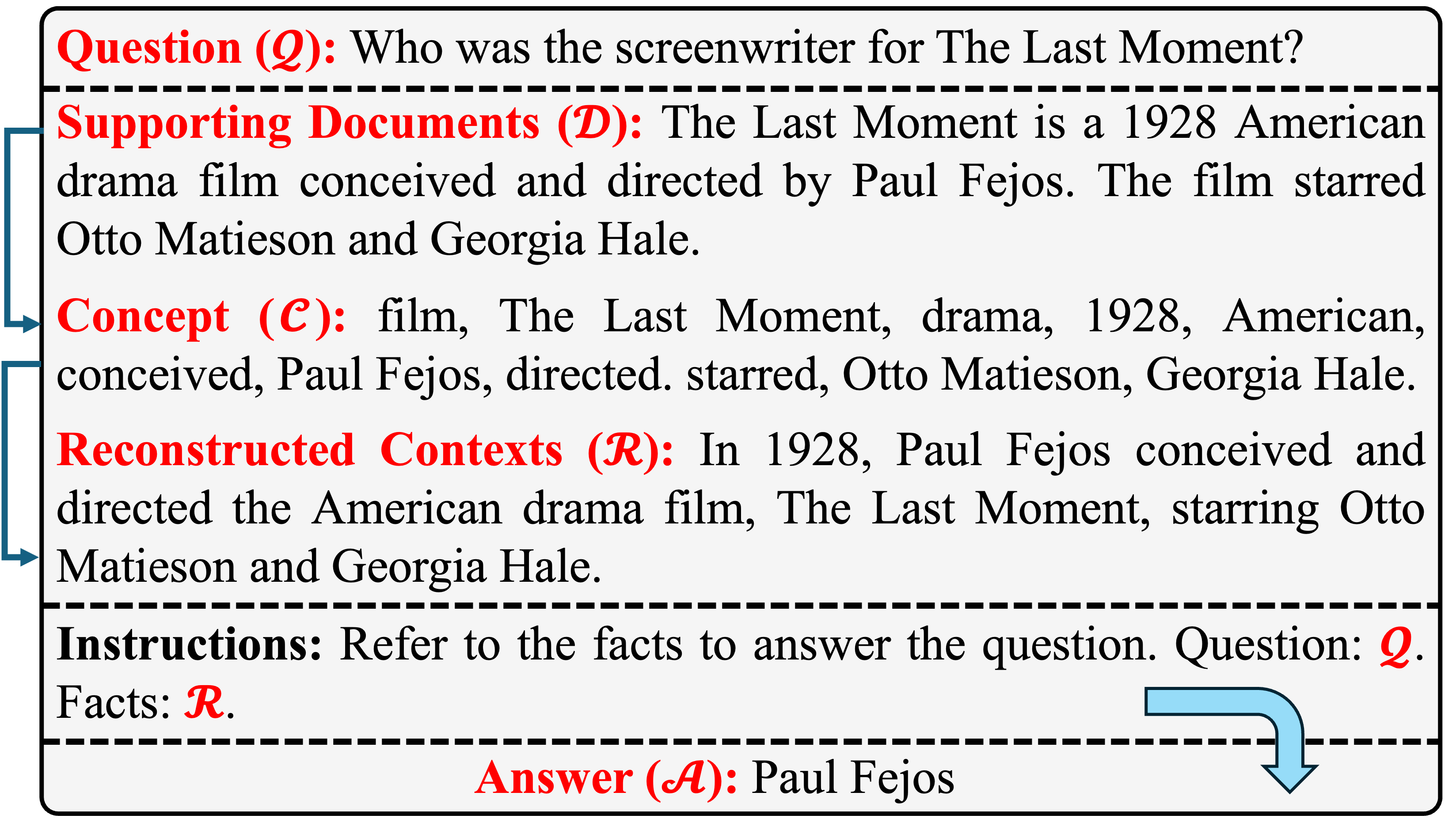}
\caption{Example of RAG based on the concept-oriented reconstructed context.}
\label{fig:top_example}
\end{figure}

Inspired by the above observations, we propose the \underline{C}oncept-\underline{o}riented \underline{C}ontext \underline{R}econstruction \underline{RAG} (CoCR-RAG) framework, which addresses the multi-source information fusion challenge in RAG by reconstructing contexts oriented to semantically grounded concepts. Fig.~\ref{fig:top_example} illustrates the processing flow of the CoCR-RAG. To capture essential concepts from complex raw documents while abstracting away potentially noisy or ambiguous details, CoCR-RAG distills the concepts using Abstract Meaning Representation (AMR)~\cite{banarescu2013abstract}, a semantic formalism that encodes the meaning of serialized texts through a rooted, directed, labeled, acyclic graph. AMR maintains the semantic consistency of concepts formatted with nodes, enabling it to normalize surface-level variations and resolve abbreviated terms in web-retrieved contexts~\cite{zhang-etal-2021-fine}. We introduce an AMR-based concept distillation algorithm that extracts and formats concepts from the structured AMR graph, producing a condensed representation of the raw multi-source contexts~\cite{liu2018amr}. This algorithm leverages the hierarchical structure of the AMR graph to identify the most salient concepts while pruning redundant information across documents from different sources. Subsequently, we instruct LLMs to fuse and reconstruct the dispersed concepts into an information-intensive context by only adding necessary sentence elements, which additionally optimizes the potentially inconsistent expressions across web sources. The reconstructed contexts are oriented to describe core concepts, serving as a unified semantic-level fused representation for the following inferences.

We evaluate the proposed CoCR-RAG in typical Web Q\&A scenarios by conducting extensive experiments on widely used datasets: PopQA~\cite{mallen-etal-2023-trust} and EntityQuestions~\cite{sciavolino-etal-2021-simple}. The results demonstrate that CoCR-RAG outperforms baselines across various context reconstruction methods and backbone LLMs. These findings highlight the effectiveness of explicitly leveraging stable linguistic features for semantic-level multi-source information fusion to enhance RAG. The contributions of this paper can be summarized as follows:

\begin{itemize}
\item We propose Concept-oriented Context Reconstruction RAG (CoCR-RAG), a framework that fuses multi-source information in Web Q\&A by supplying LLMs with concept-oriented, information-intensive reconstructed contexts. This framework achieves knowledge concentration by fusing essential information while reducing interference caused by irrelevant content.

\item We propose an AMR-based concept distillation algorithm that extracts essential concepts from multiple raw retrieved documents at the semantic level. By leveraging the structured semantic representations in AMR graphs, the algorithm enables cross-document concept-level integration, producing streamlined alternative contexts for LLMs' inference.

\item We conduct extensive experiments on the CoCR-RAG framework with widely used Web Q\&A datasets. The results show that CoCR-RAG significantly outperforms other baselines across various backbone LLMs and context reconstruction methods, demonstrating the effectiveness of stable linguistic features in supporting semantic-level fusion for refining raw contexts.
\end{itemize}

\section{Related Works}
\label{sec:related_works}

\subsection{Context Optimization}

Context optimization endeavours to enhance the efficacy of retrieval-augmented generation by refining the retrieved supporting documents to align with the requirements of the generation process. One common approach is reconstructing the context by extracting key information and removing irrelevant or redundant content. Zamani et al.~\cite{zamani2020analyzing} analyzed user interactions for search clarification and highlighted the importance of identifying and removing noisy information from the context. Li et al.~\cite{li-etal-2023-compressing} proposed the Selective Context, which compresses the context by identifying and removing redundant content to optimize the raw context. Wu et al.~\cite{wu2021context} introduced a context optimization method that leverages reinforcement learning to select the informative sentences from the retrieved documents, effectively reducing the irrelevant context while preserving the essential information.

An alternative research avenue centres on enhancing the contextual relevance of the provided query by pinpointing the most informative segments within the supporting documents. Izacard and Grave\cite{izacard2021leveraging} introduced a fusion-in-decoder architecture that allows the model to dynamically select relevant information from the context during generation, effectively focusing on the most pertinent information. In a similar vein, Laskar et al.~\cite{laskar2022query} devised a query-focused abstractive summarization technique that generates summaries of the supporting documents customized to the particular query, thus guaranteeing that the context encapsulates the utmost relevant details for addressing the query.

\subsection{Linguistics-augmented Generation}

Linguistic structures have emerged as pivotal elements in controllable text generation, enhancing semantic consistency, interpretability, precise content management, and effective handling of long-tail domains, grounded in explicit linguistic features~\cite{pmlr-v202-zhou23g}. Hardy et al.~\cite{hardy2019highres} proposed HighRES, a framework for reference-less evaluation of summarization that leverages linguistic features to assess the quality of generated summaries. Wang et al.~\cite{wang2018tree} introduced a tree-based decoder for neural machine translation that explicitly incorporates syntactic information to improve the grammaticality and coherence of the generated translations. Similarly, Li et al.~\cite{li2023explicit} presented a syntax-aware text generation approach that leverages part-of-speech tags and dependency parsing to guide the generation process. By explicitly incorporating linguistic information into the model, this approach achieves fine-grained control over the output and facilitates the analysis of the model's behavior.

Recent work has also explored using semantic representations, such as AMR, to guide the generation process~\cite{shi-etal-2023-amr}. Zhang et al.~\cite{zhang2021fine} utilized AMR to extract fine-grained information from biomedical literature, demonstrating the effectiveness of AMR in capturing domain-specific knowledge. Liu et al.~\cite{liu2020incomplete} treated incomplete utterance rewriting as a semantic segmentation task, leveraging AMR to guide the rewriting process and improve the coherence of the generated outputs. Ribeiro et al.~\cite{ribeiro2020investigating} achieved more precise control over the generated content and proposed a framework for controlling the generation process using AMR-based semantic graphs. Song et al.~\cite{song2019semantic} introduced a graph-to-text generation approach that leverages AMR to improve the faithfulness of the generated text. Hsu et al.~\cite{hsu2023ampere} incorporated AMR information into generation-based event argument extraction models. They proposed the AMPERE method, which generates AMR-aware prefixes for each layer of the generation model, introducing AMR information and improving generation quality.

\subsection{RAG in Web-based Applications}

RAG is a novel paradigm in web-based applications, particularly given its intrinsic alignment with the Web Q\&A. Recent studies have explored various aspects of RAG in the web context, including efficient retrieval mechanisms on LLMs performance~\cite{Cuconasu2024Retrieval}, dynamic update strategies for web-based knowledge bases~\cite{fan2024survey}, and context-aware re-ranking for improved relevance~\cite{xu2024list}. Additionally, researchers have investigated the integration of multimodal information from web pages, combining text, images, and structured data to enhance the comprehensiveness of retrieved information~\cite{chen-etal-2022-murag}.

Despite these advancements, a primary challenge in introducing RAG to web-based scenarios lies in the complex and heterogeneous nature of web corpora, which can significantly interfere with LLMs' inference capabilities~\cite{chen2024benchmarking}. The heterogeneity of web content, including varying writing styles and diverse formats, poses substantial difficulties in LLMs' understanding and maintaining coherence and relevance in generated outputs~\cite{shi-etal-2023-amr}. Moreover, the distracting and meaningless information surrounding key concepts across various web sources further complicates the task of reliable information retrieval and integration~\cite{yue2024evidence}. Addressing these challenges requires sophisticated filtering and verification mechanisms, along with advanced techniques for contextual understanding and cross-source information reconciliation. Linguistic elements are the basis of natural languages, which can be regarded as independent representations across various sources, presenting a viable approach for consistently capturing essential characteristics~\cite{mao2024survey}.

\section{Methods}
\label{sec:methods}

\subsection{Concept-oriented Context Reconstruction RAG Framework}
\label{sec:CoCR-RAG}

\begin{figure*}[htbp]
 \centering
 \includegraphics[width=0.95\textwidth]{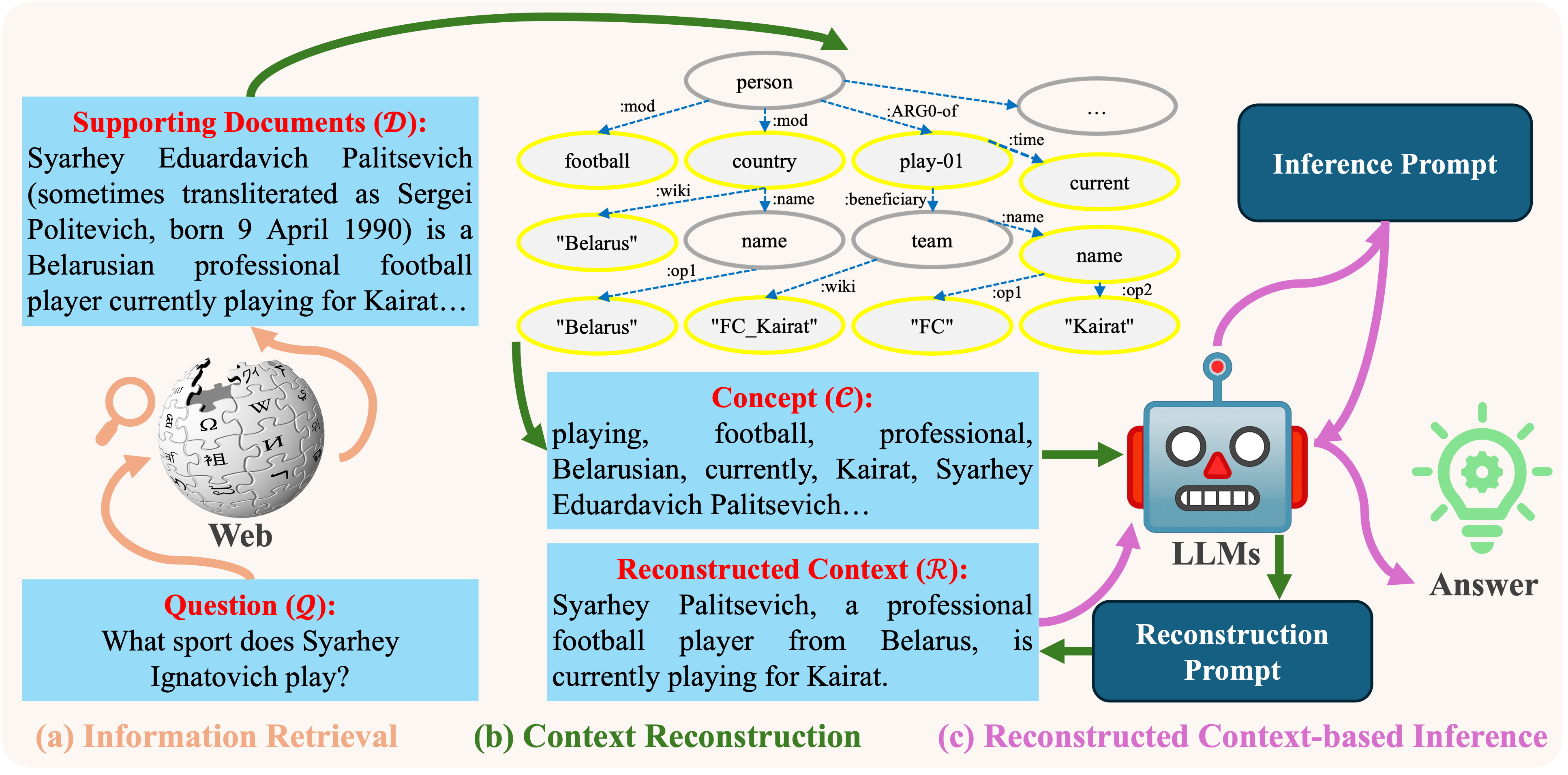}
 \caption{The overview of the Concept-oriented Context Reconstruction RAG framework, which consists of three components: \textcolor[RGB]{243, 175, 138}{(a) information retrieval}, \textcolor[RGB]{62, 127, 38}{(b) context reconstruction}, and \textcolor[RGB]{217, 116, 205}{(c) reconstructed context-based inference}. The different arrow colors indicate the corresponding components.}
 \label{fig:method_overview}
\end{figure*}

This section presents the \underline{C}oncept-\underline{o}riented \underline{C}ontext \underline{R}econstruction \underline{RAG} (CoCR-RAG) framework as Fig.~\ref{fig:method_overview}. For a given question $\mathcal{Q}$, the (a) information retrieval component aims to employ a retriever to fetch the top-$K$ informative supporting documents $\mathcal{D} = \{\mathcal{D}_1, ..., \mathcal{D}_K\}$, which are pertinent to $\mathcal{Q}$ from web sources like Wikipedia. During this phase, the efficacy of the retriever will substantially influence the answering result set $\mathcal{A} = \{a_1, ..., a_M\}$. We apply the retriever Contriever~\cite{izacard2022unsupervised} for the PopQA dataset and BM25~\cite{robertson2009probabilistic} for the EntityQuestions dataset directly, as an in-depth analysis of the retrievers is beyond the scope of this work.

The (b) context reconstruction component encompasses two critical steps: concept distillation and concept-oriented context reconstruction, which endeavors to substitute the raw supporting documents $\mathcal{D}$ by reconstructing the concept-oriented context. The concept distillation employs the AMR-based concept distillation algorithm to extract discrete concepts $\mathcal{C}$ from the raw retrieved supporting documents. Subsequently, the concept-oriented context reconstruction step leverages LLMs to reconstruct the distilled concepts back into a serialized context $\mathcal{R}$ by supplementing the basic sentence elements. These two steps reconstruct the context based on the distilled essential concepts from a linguistic perspective, thereby guiding the LLMs' inference process to focus more on the fundamental meanings conveyed by the concepts rather than on complex storylines. This approach reduces interference by filtering out irrelevant information. The details of the context reconstruction component will be expounded upon in the ensuing sections.

Upon acquiring the reconstructed context $\mathcal{R}$, the (c) reconstructed context-based inference component proceeds to amalgamate it with diverse backbone LLMs to generate responses $\mathcal{A}$ by a standardized prompt template, illustrated as: "[\textit{Refer to the facts to answer the question. Facts: $\mathcal{R}$. Question: $\mathcal{Q}$}]". Leveraging the previous hypothesis that all supporting documents contain the correct answer, our objective is to prompt LLMs to lean on the knowledge furnished by $\mathcal{R}$ when responding to $\mathcal{Q}$, reducing conflicts with their ingrained parametric knowledge to the fullest extent possible~\cite{jin2024cutting}. Moreover, existing research has shown that the strength of the prompt can influence the degree to which LLMs rely on the knowledge source~\cite{wu2024faithful}. Following this idea, we denote reconstructed context as "\textit{fact}" within the instruction, offering a constrained environment for LLMs to assume the absolute accuracy of the knowledge from $\mathcal{R}$. This enhanced nonparametric information can be coupled with various LLMs in a plug-and-play fashion. The overall framework can be formulated as Eq.~\ref{eq:framework}.

\vspace{-3mm}
\begin{equation}
P(\mathcal{A}|\mathcal{Q}) = P(\mathcal{A}|\mathcal{R},\mathcal{Q})P(\mathcal{R}|\mathcal{C},\mathcal{Q})P(\mathcal{C}|\mathcal{D},\mathcal{Q})P(\mathcal{D}|\mathcal{Q}).
\label{eq:framework}
\end{equation}

\subsection{Context Reconstruction}
\label{sec:algorithm}

\subsubsection{Concept Distillation}

Abstract Meaning Representation (AMR) is a logical semantic framework adept at encapsulating the essential common sense knowledge required to represent conceptual terms within serialized texts~\cite{santana2023survey}. Given a supporting document $\mathcal{D}_k \in \mathcal{D}$, AMR parser is applied to transform $\mathcal{D}_k$ to the AMR graph, $\mathcal{G} = <\mathcal{N},\mathcal{E}>$, where $\mathcal{N}$ and $\mathcal{E}$ denote the nodes for concepts and edges for relationships. A mBart-based~\cite{liu2020multilingual} parser~\footnote{\url{https://github.com/BramVanroy/multilingual-text-to-amr}} trained on the AMR 3.0 dataset~\footnote{\url{https://catalog.ldc.upenn.edu/LDC2020T02}} is introduced for parsing to address potential multilingual concerns in Web-based content. Table~\ref{tab:AMR_tc} shows an example of parsing process.

\begin{algorithm}
\caption{Concept-oriented Context Reconstruction}
\label{alg:c_d}
\SetAlgoLined
\SetKwInOut{Input}{Input}
\SetKwInOut{Output}{Output}
\SetKwInOut{Repeat}{repeat}
\Input{AMR Graph ($\mathcal{G}$)}
\Output{concept ($\mathcal{C}$)}
\SetKwFunction{SplitSnt}{SplitSnt}
\SetKwFunction{AppendConcept}{AppendConcept}
\SetKwFunction{SearchRole}{SearchRole}
\SetKwFunction{AppendRole}{AppendRole}
\SetKwFunction{HandleRole}{HandleRole}
\SetKwFunction{IsRole}{IsRole}
\SetKwFunction{ConceptBacktrace}{ConceptBacktrace}
\SetKwFunction{ConceptFormat}{ConceptFormat}
\SetKwFunction{IsWiki}{IsWiki}
\SetKwFunction{HandleWiki}{HandleWiki}
\SetKwFunction{IsDate}{IsDate}
\SetKwFunction{HandleDate}{HandleDate}
\SetKwFunction{IsName}{IsName}
\SetKwFunction{HandleName}{HandleName}
\SetKwFunction{Recon}{Recon}
\SetKwFunction{DFS}{DFS}
\SetKwProg{Fn}{Function}{:}{}
\Fn{Concept\_Distillation\textup{($\mathcal{G}$)}}{
    concept $\gets []$, role $\gets []$\;
    \For{\textup{snt} in \SplitSnt \textup{($\mathcal{G}$)}}{
        \For{$\mathcal{N}$ in \DFS{\textup{snt}}}{
            \uIf{\IsRole{$\mathcal{N}$}}{
                \uIf{\IsName{$\mathcal{N}$}}{
                    \AppendRole{\HandleName{$\mathcal{N}$}}
                }
                \uIf{\IsWiki{$\mathcal{N}$}}{
                    \AppendRole{\HandleWiki{$\mathcal{N}$}}
                }
                \uIf{\IsDate{$\mathcal{N}$}}{
                    \AppendRole{\HandleDate{$\mathcal{N}$}}
                }
            }
            \uElse{
                \uIf{\textup{role} is not None}{
                    \AppendConcept{\HandleRole{\textup{role}}}\;
                    role $\gets []$\;
                }
                \AppendConcept{$\mathcal{N}$}\;
            }
            \uIf{($\mathcal{N}$is Last\textup{)} and \textup{(role} is not \textup{None)}}{
            \Repeat{Algorithm.Line 5-11}
            \AppendConcept{\HandleRole{\textup{role}}}\;
            }
        }
    }
    concept $\gets$ \ConceptFormat(concept)\;
    concept $\gets$ \ConceptBacktrace(concept)\;
    \Return $\mathcal{C}$ $\gets$ concept
}
\Fn{Context\_Reconstruction\textup{($\mathcal{C}$)}}{
    $\mathcal{R}$ $\gets$ \Recon($\mathcal{C}$)\;
    \Return $\mathcal{R}$
}
\end{algorithm}

We introduce the concept distillation algorithm to format the concepts portrayed in $\mathcal{G}$, as delineated in Algorithm~\ref{alg:c_d}. Each sentence $snt \in \mathcal{D}_k$ can be structurally parsed into a predefined \texttt{multi-sentence} format. The \texttt{SplitSnt($\cdot$)} function is devised to segment $\mathcal{G}$ and arrange the resultant sentence-based sub-graphs based on their sequential order. Since the essential concepts are carried by $\mathcal{N}$, we simplify $\mathcal{G}$ by ignoring $\mathcal{E}$ representing the nodes' relations. Notably, the filtering of relationships between concepts is aimed at preventing the introduction of extraneous noise not explicitly present in the raw supporting documents. These relationships will be restored in context reconstruction, leveraging the basic language logic learned by the LLMs during pre-training. Consequently, $\mathcal{G}$ is streamlined into a unidirectional connecting structure. We execute a Depth First Search, \texttt{DFS($\cdot$)}, for traversing $\mathcal{N}$ of $\mathcal{G}$ to format the concepts for preserving the relative position of adjacent concept-carrying nodes. This traversal setting will reduce the potential interference of the random distribution of concepts in the context reconstruction process. This traversal process is elucidated in Fig.~\ref{fig:AMR_DFS}.

The AMR delineates a set of roles to define the semantic structure of sentences meticulously. This algorithm specifically emphasizes three roles: \texttt{:name}, \texttt{:wiki}, and \texttt{date-entity}, utilizing \texttt{IsRole($\cdot$)} for identification. The \texttt{:name} role represents the nodes to carry the entities. The AMR parsing process will decompose each word in \texttt{:name} into a predicate role (\texttt{:op}) and disperse the overall concept to different nodes when the concept expressed by \texttt{:name} consists of multiple words. The dispersed nodes could confuse the \texttt{DFS($\cdot$)} process, which will interfere with the LLMs for the following context reconstruction. To preserve the integrity of concepts expressed by \texttt{:name}, we introduce \texttt{HandleName($\cdot$)}, organizing predicates in a stack structure to avoid confusion caused by independent understanding. For introducing reliable external knowledge references, the \texttt{:wiki} role is processed to standardize diverse expressions referring to the same concept through \texttt{HandleWiki($\cdot$)}, aligning them with the corresponding definitions sourced from Wikipedia. If the concept expressed by a node differs between \texttt{:name} and \texttt{:wiki}, we designate the concept as \texttt{:wiki} to avoid semantic ambiguity by the standard knowledge base. Moreover, the \texttt{date-entity} role represents temporal concepts. In this algorithm, we handle the roles \texttt{:year}, \texttt{:month}, and \texttt{:day} using \texttt{HandleDate($\cdot$)}. This function consolidates roles under the same \texttt{date-entity}, forming the numerical month concept to its textual representations for more discriminative formatting. AMR incorporates special numerical instructions for certain parsed nodes, such as \texttt{play-01} in Fig.~\ref{fig:method_overview}, where the word-following numbers indicate various meanings of the same word in different contexts as defined in OntoNotes. In the RAG scenario, LLMs are presented with supporting documents comprising multiple concepts, implying that concepts are understood in conjunction with relevant linguistic contexts rather than independently. Therefore, our algorithm relies on the intrinsic contextual reasoning ability of LLMs to distinguish and understand nuanced polysemous concepts rather than providing explicit complex and redundant semantic references that would instead introduce noise. The function \texttt{HandleRole($\cdot$)} formats the processed roles to the preliminary concept set, while the nodes not falling within the above roles are processed directly by their instance with \texttt{AppendConcept($\cdot$)}. 

The structure of AMR includes a collection of pre-defined nodes (\texttt{government-organization}, \texttt{country-region}, etc.) to enforce knowledge and prevent hallucination regarding entity types during AMR-based generation. However, the inference process of LLMs in our scenario is not directly based on the AMR graph but on the contexts reconstructed by the formatted concepts. The LLMs can self-organize and understand discrete concepts relying on their memorized linguistic knowledge in the context reconstruction process, while the pre-defined nodes will significantly interfere with this process by introducing redundant noise. To address this concern, we employ \texttt{ConceptFormat($\cdot$)} for filtering out these nodes. Additionally, we eliminate frequently occurring concepts in the formulation process to maximize the elimination of interference, as these concepts tend to dilute the information density for reconstruction. 

Furthermore, AMR adopts the principle of abstraction and generalization in its representation, rather than using exact lexical items. This approach to representation risks leading the nodes to overlook important contextual variations such as tense. To ensure consistency with the concepts conveyed in the original supporting documents, we introduce the \texttt{Backtrace($\cdot$)} function. By inducting the formatted concepts into the corresponding representation in $\mathcal{D}$, this function facilitates the alignment of the generated representation with the originally expressed semantics. The backtraced concepts serve as the finalized concept $\mathcal{C}$, forming the basic semantic element for constraining the context reconstruction.

\subsubsection{Concept-oriented Context Reconstruction}

We reconstruct the context by \texttt{Recon($\cdot$)} in Algorithm~\ref{alg:c_d} based on previously distilled concepts $\mathcal{C}$. This strategy optimizes the raw context representation with conceptual constraining, allowing it to center on pertinent information while attenuating extraneous details. Specifically, we introduce LLMs for concept-oriented context reconstruction, employing the prompt template elucidated in Prompt~\ref{prompt:resc} following the format defined by Taori et al~\cite{alpaca}. For the concepts in each document, we instruct the LLMs to construct succinct sentences to encapsulate all relevant concepts, which are described as "\texttt{keywords}" to intuitively provide the linguistic-only description. This process instructs LLMs to incorporate only essential linguistic elements for sentence-making, eliminating superfluous and concept-irrelevant words to avoid overthinking. This procedure is model-agnostic and can be adapted across various LLM architectures. In this study, we specifically prompt the LLaMA-2-13b-chat-hf~\cite{touvron2023llama} to execute instructions. The reconstructed concept-oriented context is denoted as $\mathcal{R}$, the optimized context for inference.

\begin{tcolorbox}[
    colframe=black!75, 
    label=prompt:resc, 
    fonttitle=\bfseries, 
    title={Prompt 1: The prompt for instructing LLMs in concept-oriented context reconstruction.}, before upper=\refstepcounter{mytcolorbox}\label{prompt:resc}
]
{[INST] <<SYS>> \\ Instruction: \texttt{Make short sentences containing all the following keywords by adding the necessary sentence elements only.}\\ 
\text{<</SYS>>}} \\
{Prompt = """Below is an instruction that describes a task, paired with an input that provides keywords.\\     \#\#\# Instruction: \{""" + \texttt{Instruction} + """\}\\    \#\#\# Input: \{""" + $\mathcal{C}$ + """\}\\     \#\#\# Response: """} \\
\text{[/INST]} 
\end{tcolorbox}

\section{Experiments}
\label{sec:experiments}

\subsection{Datasets}
\label{sec:datasets}

To verify the effectiveness of the proposed CoCR-RAG, we perform evaluations on two question-answering datasets: PopQA~\cite{mallen-etal-2023-trust} and EntityQuestions~\cite{sciavolino-etal-2021-simple}. These datasets consist of supporting documents retrieved from Wikipedia, a typical source for Web Q\&A. We investigate the parameter $K$, which denotes the number of supporting documents associated with each question $\mathcal{Q}$. As $K$ increases, the number of supporting documents grows, resulting in a corresponding increase in overall context length. The statistical results regarding the number of selected pairs with different $K$ are in Table~\ref{tab:datasets}.

\begin{table}[htbp]
\centering
\caption{Statistical results for the number of "<$\mathcal{Q}$-$\mathcal{A}$-$\mathcal{R}$>" pairs screened out from the datasets.}
\begin{adjustbox}{width=1\linewidth}
\begin{tabular}{c|c|c|c|c|c|c|c|c|c|c}
\hline
$K$= & 1 & 2 & 3 & 4 & 5 & 6 & 7 & 8 & 9 & 10 \\ \hline \hline
PopQA~\cite{mallen-etal-2023-trust} & 3353 & 2339 & 862 & 468 & 258 & 155 & 72 & 37 & 19 & 6 \\ \hline
EntityQuestions~\cite{sciavolino-etal-2021-simple} & 1766 & 1048 & 632 & 417 & 292 & 224 & 163 & 91 & 68 & 24 \\
\hline
\end{tabular}
\end{adjustbox}
\label{tab:datasets}
\end{table}

\subsection{Baselines}
\label{sec:baselines}

The baseline evaluations cover two aspects: (1) evaluating CoCR-RAG's adaptability across various backbone LLMs, and (2) comparing CoCR-RAG with diverse context reconstruction techniques. We use publicly available LLMs as backbones, including GPT-Neo-1.3B, GPT-Neo-2.7B~\cite{gpt-neo}, GPT-j-6b~\cite{gpt-j}, OPT-1.3b, OPT-2.7b, OPT-6.7b~\cite{zhang2022opt}, bloom-560m, bloom-1b1, bloom-1b7, bloom-3b~\cite{le2022bloom}, LLaMA-2-7b-chat-hf, LLaMA-2-13b-chat-hf~\cite{touvron2023llama}. We encourage LLMs to rely solely on the knowledge in supporting documents. Commercial LLMs are excluded since their frequent updating is likely to parameterize testing knowledge. The backbone LLMs that are coupled with raw supporting documents are the Vanilla methods. CoCR-RAG reconstructs the context into information-intensive representations by reducing irrelevant information. We adopt keyword extraction, context summarization, Selective Context (SelCon)~\cite{li-etal-2023-compressing}, and LLMLingua~\cite{jiang-etal-2023-llmlingua} as the baseline reconstruction techniques, which also aim to improve inference efficiency by reducing interference information in contexts.

Drawing inspiration from Chuang et al.~\cite{chuang2024learning} and Laskar et al.~\cite{laskar2022query}, we instruct LLaMA-2-13b-chat-hf~\cite{touvron2023llama} to establish baselines of context keyword extraction and summarization. This setting involves extracting key elements from raw supporting documents to generate concise context representations, guided by prompts of "[\texttt{Extract a few keywords from the following content.}]" and "[\texttt{Generate a short summary of the following content.}]". The complete prompts are detailed in Prompt~\ref{prompt:baselines}. Additionally, the SelCon baseline incorporates a hyperparameter, the reduction ratio, to control the proportion of content to be filtered, which is set to $0.5$ in this comparison. For the LLMLingua baseline, the hyperparameter controlling the expected context length is set to $200$, which is the default value. These foundational configurations highlight the advantages of CoCR-RAG from both prompt-based and linguistic-based perspectives.

\subsection{Evaluation Metrics}
\label{sec:evaluation}

We employ two distinct metrics to evaluate the performance of CoCR-RAG: accuracy ($Acc$) and Area Under the Curve ($AUC$). The accuracy metric ($Acc$) is calculated following the methodology outlined by Mallen et al.~\cite{mallen-etal-2023-trust}, measuring whether any substring of the generated answers $\mathcal{A}$ exactly matches any of the gold-standard answers when provided with $\mathcal{Q}$. On the other hand, the $AUC$ metric is designed to offer a broader and more nuanced evaluation of the model's performance across a varying number of supporting documents $K$. The computation of $AUC$ is carried out according to Eq.~\ref{eq:Inte}, where $x_s$ and $x_e$ denote the endpoints of the interval where $K$ lies. Higher $AUC$ values indicate superior overall performance across a range of $K$ values, highlighting the model's overall performance.

\begin{equation}
\begin{aligned}
    AUC &= \int_{x_s}^{x_e} Acc(x) \, dx \\
    &\approx \frac{1}{2} \sum_{i=1}^{K} (x_{i} - x_{i-1}) 
    \left[ Acc(x_{i}) + Acc(x_{i-1}) \right]
\end{aligned}
\label{eq:Inte}
\end{equation}

\section{Results and Analysis}
\label{sec:results}

\begin{figure*}[htbp]
 \centering
 \includegraphics[width=1\textwidth]{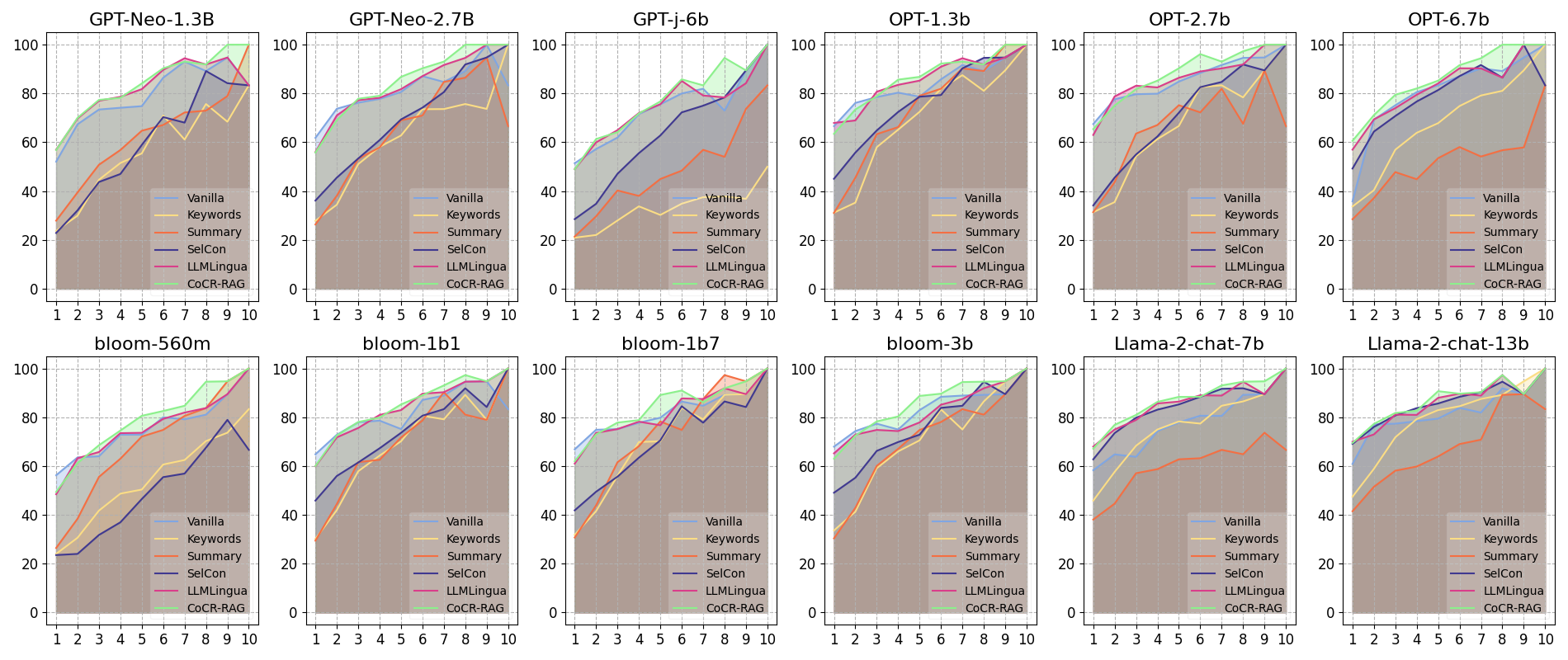}
 \caption{The evaluation results depict trends in $Acc. \uparrow$ and $AUC\uparrow$ on the PopQA dataset. The vertical axis denotes $Acc$, while the horizontal axis represents the number of supporting documents, $K$. The polyline illustrates the fluctuation in $Acc$ with varying $K$. The shaded area represents $AUC$}
 \label{fig:PopQA}
\end{figure*}

\begin{figure*}[htbp]
 \centering
 \includegraphics[width=1\textwidth]{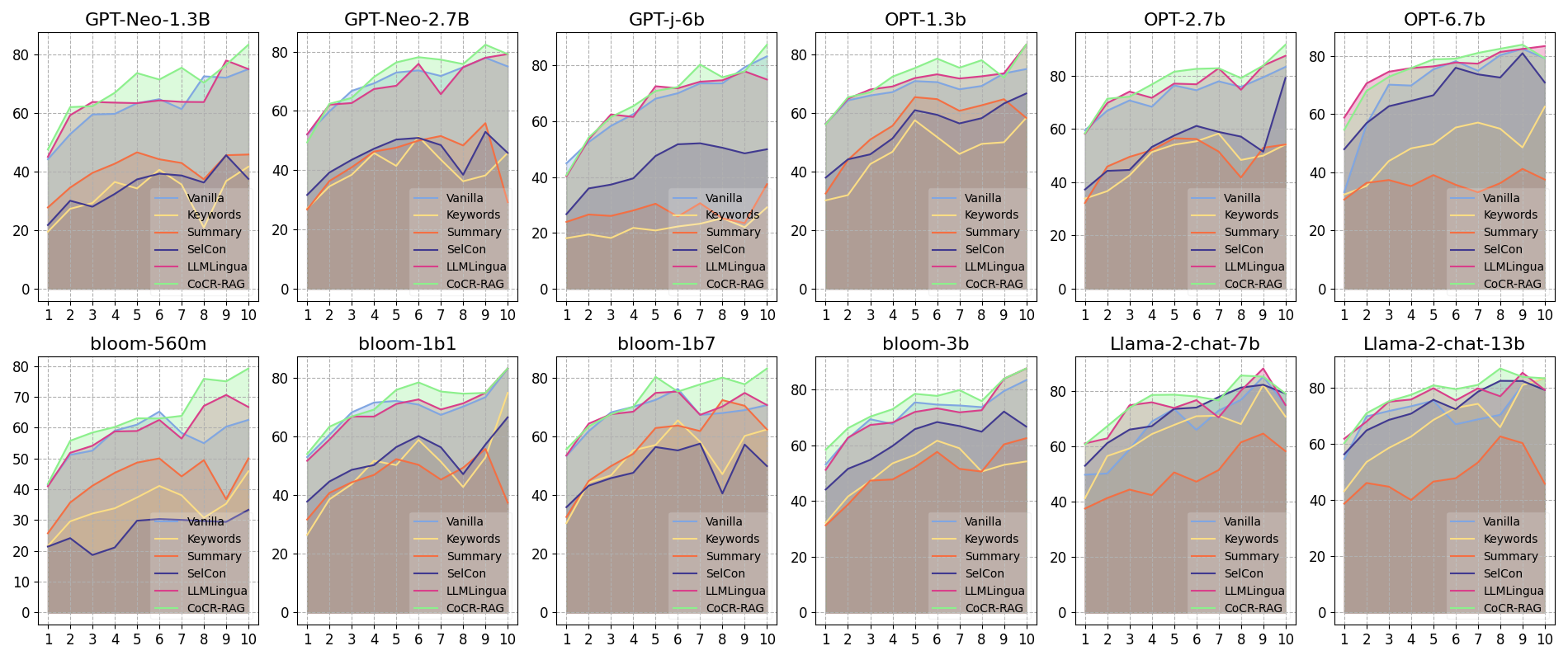}
 \caption{The evaluation results illustrate trends in $Acc. \uparrow$ and $AUC\uparrow$ on the EntityQuestion dataset, following the same axis and symbol definitions as in Fig.~\ref{fig:PopQA}.}
 \label{fig:EQ}
\end{figure*}

\begin{table*}[htbp]
\centering
\caption{The quantitative results of $AUC\uparrow$ for the PopQA dataset, with the full name order of the LLMs as follows: GPT-Neo-1.3B, GPT-Neo-2.7B, GPT-j-6b, OPT-1.3b, OPT-2.7b, OPT-6.7b, bloom-560m, bloom-1b1, bloom-1b7, bloom-3b, LLaMA-2-chat-7b, LLaMA-2-chat-13b. The best results are in \textbf{bold}, and the next best results are in \uline{underlined}. The \textcolor{applegreen}{increased} $\Delta$ has been marked.}
\setlength{\tabcolsep}{3pt}
\begin{tabular}{cccccccccccccc}
\hline
$\mathcal{D}$ & G-1.3 & G-2.7 & G-j-6 & O-1.3 & O-2.7 & O-6.7 & b-560 & b-1b1 & b-1b7 & b-3 & L-7 & L-13 & $\sigma$\\ \hline \hline
{\rotatebox{0}{{Vanilla}}} & 539.91 & {555.70} & {490.49} & {569.08} & {583.23} & {548.19} & {501.18} & {560.34} & \uline{556.89} & \uline{565.54} & 515.85 & 554.59 & \uline{27} \\ \hline
{\rotatebox{0}{{Keywords}}} & 363.29 & 405.28 & 215.69 & 456.49 & 438.54 & 440.44 & 341.82 & 453.85 & 461.18 & 455.32 & 507.55 & 533.24 & 81 \\ \hline
{\rotatebox{0}{{Summary}}} & 401.59 & 430.34 & 295.96 & 486.48 & 453.17 & 338.15 & 439.64 & 464.63 & 478.31 & 462.03 & 404.63 & 438.62 & 55 \\ \hline
{\rotatebox{0}{{SelCon}}} & 376.17 & 447.93 & 400.95 & 511.20 & 465.31 & 539.90 & 297.28 & 491.25 & 465.67 & 504.65 & {579.12} & \uline{586.74} & 80 \\ \hline
{\rotatebox{0}{{LLMLingua}}} & \uline{565.77} & \uline{562.33} & \uline{500.55} & \uline{583.85} & \uline{587.63} & \uline{559.34} & \uline{503.56} & \uline{568.57} & 555.43 & 551.21 & \uline{585.21} & 585.25  & 28\\ \hline
{\rotatebox{0}{\cellcolor{intercolor_1}{\textbf{CoCR-RAG}}}} & \cellcolor{intercolor_1}\textbf{567.45} & \cellcolor{intercolor_1}\textbf{574.05} & \cellcolor{intercolor_1}\textbf{514.95} & \cellcolor{intercolor_1}\textbf{587.84} & \cellcolor{intercolor_1}\textbf{602.97} & \cellcolor{intercolor_1}\textbf{584.28} & \cellcolor{intercolor_1}\textbf{524.32} & \cellcolor{intercolor_1}\textbf{576.56} & \cellcolor{intercolor_1}\textbf{572.85} & \cellcolor{intercolor_1}\textbf{582.81} & \cellcolor{intercolor_1}\textbf{594.88} & \cellcolor{intercolor_1}\textbf{595.84} & \cellcolor{intercolor_1}\textbf{26} \\ \hline \hline
{\rotatebox{0}{{$\Delta$}}} & \textcolor{applegreen}{+27.54} & \textcolor{applegreen}{+18.35} & \textcolor{applegreen}{+24.46} & \textcolor{applegreen}{+18.76} & \textcolor{applegreen}{+19.74} & \textcolor{applegreen}{+36.09} & \textcolor{applegreen}{+23.14} & \textcolor{applegreen}{+16.22} & \textcolor{applegreen}{+15.96} & \textcolor{applegreen}{+17.27} & \textcolor{applegreen}{+79.03} & \textcolor{applegreen}{+41.25} & - \\ \hline
\end{tabular}
\label{tab:PopQA}
\end{table*}

\begin{table*}[htbp]
\centering
\caption{The quantitative results of $AUC\uparrow$ for the EntityQuestions dataset follow the LLMs' order and symbol definitions as presented in Table~\ref{tab:PopQA}.}
\setlength{\tabcolsep}{3pt}
\begin{tabular}{cccccccccccccc}
\hline
$\mathcal{D}$ & G-1.3 & G-2.7 & G-j-6 & O-1.3 & O-2.7 & O-6.7 & b-560 & b-1b1 & b-1b7 & b-3 & L-7 & L-13 & $\sigma$ \\ \hline \hline
{\rotatebox{0}{{Vanilla}}} & {419.76} & \uline{477.67} & {444.72} & {469.79} & {502.47} & {481.28} & {395.26} & \uline{473.16} & {477.78} & \uline{486.80} & 454.46 & 488.31 & \textbf{30} \\ \hline
{\rotatebox{0}{{Keywords}}} & 223.65 & 286.79 & 147.71 & 316.44 & 339.38 & 333.05 & 238.12 & 329.76 & 366.60 & 360.61 & 445.11 & 444.69 & 83 \\ \hline
{\rotatebox{0}{{Summary}}} & 283.05 & 310.28 & 192.44 & 389.03 & 348.75 & 250.26 & 302.53 & 321.16 & 390.52 & 335.85 & 327.42 & 329.31 & 53 \\ \hline
{\rotatebox{0}{{SelCon}}} & 234.72 & 314.73 & 303.06 & 366.26 & 366.92 & 460.20 & 179.64 & 359.63 & 344.78 & 421.39 & {487.63} & {499.92} & 93 \\ \hline
{\rotatebox{0}{{LLMLingua}}} & \uline{432.41} & 465.59 & \uline{453.87} & \uline{483.31} & \uline{518.63} & 522.23 & \uline{396.49} & 467.85 & \uline{480.73} & 476.79 & \uline{505.94} & \uline{523.09} & 36\\ \hline
{\rotatebox{0}{\cellcolor{intercolor_1}{\textbf{CoCR-RAG}}}} & \cellcolor{intercolor_1}\textbf{470.74} & \cellcolor{intercolor_1}\textbf{492.66} & \cellcolor{intercolor_1}\textbf{463.07} & \cellcolor{intercolor_1}\textbf{501.31} & \cellcolor{intercolor_1}\textbf{535.93} & \cellcolor{intercolor_1}\textbf{523.77} & \cellcolor{intercolor_1}\textbf{422.84} & \cellcolor{intercolor_1}\textbf{494.17} & \cellcolor{intercolor_1}\textbf{503.37} & \cellcolor{intercolor_1}\textbf{512.23} & \cellcolor{intercolor_1}\textbf{526.92} & \cellcolor{intercolor_1}\textbf{538.31} & 
\cellcolor{intercolor_1}\uline{32}\\ \hline \hline
{\rotatebox{0}{{$\Delta$}}} & \textcolor{applegreen}{+50.98} & \textcolor{applegreen}{+14.99} & \textcolor{applegreen}{+18.35} & \textcolor{applegreen}{+31.52} & \textcolor{applegreen}{+33.46} & \textcolor{applegreen}{+42.49} & \textcolor{applegreen}{+27.58} & \textcolor{applegreen}{+21.01} & \textcolor{applegreen}{+25.59} & \textcolor{applegreen}{+25.43} & \textcolor{applegreen}{+72.46} & \textcolor{applegreen}{+50.00} & -\\ \hline
\end{tabular}
\label{tab:EQQA}
\end{table*}

The evaluation results of $Acc$ with various $K$ values in the PopQA and EntityQuestion datasets are respectively shown in Fig.~\ref{fig:PopQA} and Fig.~\ref{fig:EQ}, providing an intuitive graphical trend as $K$ increases. Moreover, the quantitative results of $AUC$ for both datasets are in Table~\ref{tab:PopQA} and Table~\ref{tab:EQQA}. Since the sample sizes in the two datasets are small when $K=9$ and $K=10$, $AUC$ in these tables is calculated based on the interval of $K \in [1,8]$. The tables incorporate the calculation of $\Delta$, which quantifies the enhancement attained by the CoCR-RAG compared to the Vanilla baselines. Specifically, $\Delta$ is calculated as follows: $\Delta={AUC.}_{CoCR}-{AUC.}_{Vanilla}$. Table~\ref{tab:aca_PopQA} and Table~\ref{tab:aca_EQQA} provide the full results of $Acc$ across $K \in [1,10]$.

The comparison depicted in Fig.~\ref{fig:PopQA} and Fig.~\ref{fig:EQ} underscores the notable efficacy of the proposed CoCR-RAG, particularly evident as $K$ increases. The CoCR-RAG outperforms most of the other baselines in both datasets when $K \in [4, 8]$, a balanced interval with a lengthy context and substantial samples. This finding demonstrates the effectiveness of CoCR-RAG in mitigating the impact of irrelevant information in complex context scenarios. When $K=1$, the gains from all reconstruction methods compared to the Vanilla baseline are minimal since the entropy here is relatively stable. At this point, the information contained in the context approaches an optimal state of information transmission that can support the inference well. Any additional operations may introduce extraneous noise, reducing the proportion of effective information. This trend suggests that context reconstruction should assess whether the entropy is sufficient to support downstream inference.

All the $\Delta$ in Table~\ref{tab:PopQA} and Table~\ref{tab:EQQA} are positive, generally proving the significance of providing the alternative informative reconstructed context rather than the messy raw supporting documents. Simultaneously, it can be observed that for models with inherently longer context windows, such as LLaMA-2-chat-7b/13b, the improvement in $\Delta$ is more pronounced. This finding emphasizes that LLMs with longer context windows are more susceptible to incorporating noisy information due to their enhanced capacity for embedding richer information. This makes the performance improvement of concept-oriented context reconstruction most prominent, proving that it is as crucial as expanding the context windows.

The baseline configurations can be divided into two technical approaches: lexical-focused (Keywords, SelCon) and syntactic-focused (Summary, LLMLingua), which reconstruct the context into discrete words or more understandable sentences, respectively. LLMLingua focuses on finer-grained pruning within the characters of words, allowing it to retain a relatively complete syntactic structure after compression that can be regarded as a syntactic-focused approach. Lexical-focused methods offer more efficient filtering of irrelevant information, but discrete representations may hinder small-scale LLMs comprehension due to their insufficient parameterized knowledge. Syntactic-focused methods enhance the interpretability of the reconstructed context as they keep the basic language structures and exhibit greater robustness in inferences, but they may retain redundant information. These characteristics are evident when comparing Keywords with Summary and SelCon with LLMLingua. Lexical-focused methods perform competitively with large inference LLMs, while syntactic-focused methods show a clear advantage with smaller LLMs and show lower overall standard deviations ($\sigma$), indicating more significant robustness gains for downstream inferences. CoCR-RAG integrates the strengths of both technical approaches, with its lexical-focused concept distillation screening the essential information in a minimal granular, while the concept-oriented reconstruction enhances interpretability by supplementing concepts with basic language elements. These complementary advantages enable CoCR-RAG to achieve the highest $AUC$ performance while maintaining the lowest $\sigma$ among baselines, demonstrating it is a flexible plug-and-play augmentation module that can be coupled with various backbone LLMs.

In terms of applicability, CoCR-RAG’s concept distillation process is based on the inherent linguistic patterns of natural language, eliminating the need for hyperparameter tuning and avoiding the complexity of such settings in practical applications. In contrast, SelCon and LLMLingua require hyperparameter tuning to determine the optimal granularity of redundancy elimination, potentially complicating the application process. Additionally, compared to fully instruction-guided context reconstruction baselines of Keywords and Summary, CoCR-RAG provides additional linguistic constraints during the reconstruction inference phase. Such constraints prevent inference biases that arise from purely abstract instructions.

\section{Conclusions and Future Research}
\label{sec:conclusion}

This paper proposes the CoCR-RAG framework to address multi-source information fusion in RAG-based Web Q\&A. The framework uses AMR to distill essential concepts from raw documents and reconstructs them into unified contexts via LLMs. These reconstructed contexts provide a semantically fused representation, allowing LLMs to focus on relevant knowledge within noisy web data. Experiments on PopQA and EntityQuestions show that CoCR-RAG consistently outperforms baseline methods and remains robust across different backbone LLMs, demonstrating the effectiveness of linguistic-based context reconstruction as a plug-and-play enhancement for RAG.

The current reconstruction process relies on instruction-guided inference and will be affected by parametric uncertainty. Future work may explore stable linguistic approaches to improve generalization, as well as alternative structures such as dependency parsing and semantic role labeling. Moreover, this study shows that linguistically grounded representations offer a practical foundation for information fusion and can be extended to other text-intensive tasks.

\section*{Acknowledgments}
\label{sec:ack}

This research is supported by the Australian Research Council (ARC) Under Grants DP220103717 and LE220100078, and the National Natural Science Foundation of China under Grants No.62072257. 


\bibliographystyle{cas-model2-names}

\bibliography{cas-refs}

\appendix

\section*{Appendix}

\setcounter{figure}{0}
\setcounter{table}{0}
\setcounter{mytcolorbox}{0}
\renewcommand{\thefigure}{A\arabic{figure}}
\renewcommand{\thetable}{A\arabic{table}}
\renewcommand{\themytcolorbox}{A\arabic{mytcolorbox}}

\section{Qualitative Examples}
\label{sec:AMR_example}

\begin{figure*}[!htbp]
 \centering
 \includegraphics[width=1\textwidth]{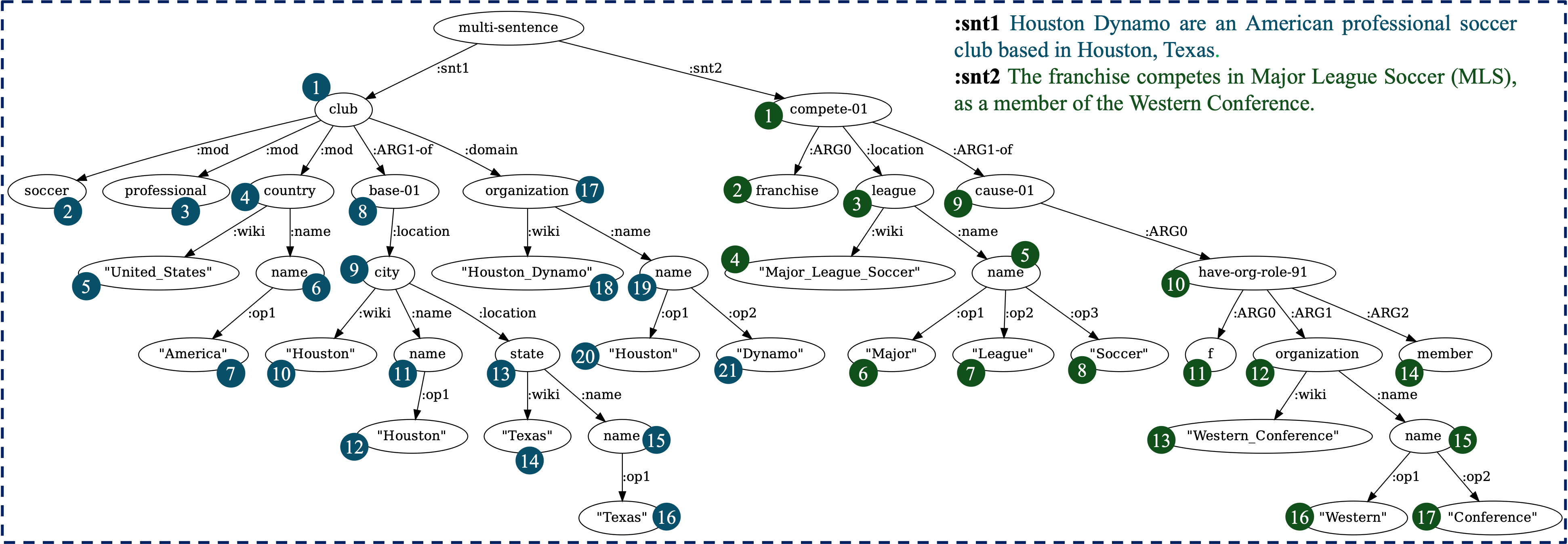}
 \caption{The node order traversed by \texttt{DFS($\cdot$)} in the AMR graph, where the nodes marked here are not the final distilled concepts.}
 \label{fig:AMR_DFS}
\end{figure*}

An AMR graph parsed from a raw supporting document in the PopQA dataset, "\texttt{Houston Dynamo are an American professional soccer club based in Houston, Texas. The franchise competes in Major League Soccer (MLS), as a member of the Western Conference}", is used as an example in this section. The \texttt{DFS($\cdot$)} traversal order of the graph nodes is illustrated in Fig.~\ref{fig:AMR_DFS}. The parsed AMR graph, annotated with the distilled concepts, is presented in Table~\ref{tab:AMR_tc}, which demonstrates that the \texttt{DFS($\cdot$)} traversal maintains the relative order of adjacency concepts.

\begin{table}[!htbp]
\centering
\caption{An example of the parsed AMR graph. Nodes containing concepts are \colorbox{rolecolor_2}{highlighted}, including "club, soccer, professional, American, based, Texas, Houston Dynamo. competes, franchise, Major League Soccer, Western Conference, member".}
\label{tab:AMR_tc}
\begin{tabular}{p{1\linewidth}}
\hline
\textbf{Supporting Document:} Houston Dynamo are an American professional soccer club based in Houston, Texas. The franchise competes in Major League Soccer (MLS), as a member of the Western Conference. \\
\hline
\hspace{1cm}(m / multi-sentence \\
\hspace{1.5cm}:snt1 (c / \cellcolor{rolecolor_2}{club} \\
\hspace{2cm}:mod (s / \cellcolor{rolecolor_2}{soccer}) \\
\hspace{2cm}:mod (p / \cellcolor{rolecolor_2}{professional}) \\
\hspace{2cm}:mod (c2 / country \\
\hspace{2.5cm}:wiki "United\_States" \\
\hspace{2.5cm}:name (n / name \\
\hspace{3cm}:op1 \cellcolor{rolecolor_2}{"America"})) \\
\hspace{2cm}:ARG1-of (b / \cellcolor{rolecolor_2}{base-01} \\
\hspace{2.5cm}:location (c3 / city \\
\hspace{3cm}:wiki \cellcolor{rolecolor_2}{"Houston"} \\
\hspace{3cm}:name (n2 / name \\
\hspace{3.5cm}:op1 \cellcolor{rolecolor_2}{"Houston"}) \\
\hspace{3cm}:location (s2 / state \\
\hspace{3.5cm}:wiki \cellcolor{rolecolor_2}{"Texas"} \\
\hspace{3.5cm}:name (n3 / name \\
\hspace{4cm}:op1 \cellcolor{rolecolor_2}{"Texas"}))))) \\
\hspace{2cm}:domain (o / organization \\
\hspace{2.5cm}:wiki \cellcolor{rolecolor_2}{"Houston\_Dynamo"} \\
\hspace{2.5cm}:name (n4 / name \\
\hspace{3cm}:op1 \cellcolor{rolecolor_2}{"Houston"} \\
\hspace{3cm}:op2 \cellcolor{rolecolor_2}{"Dynamo"}))) \\
\hspace{1.5cm}:snt2 (c4 / \cellcolor{rolecolor_2}{compete-01} \\
\hspace{2cm}:ARG0 (f / \cellcolor{rolecolor_2}{franchise}) \\
\hspace{2cm}:location (l / league \\
\hspace{2.5cm}:wiki \cellcolor{rolecolor_2}{"Major\_League\_Soccer"} \\
\hspace{2.5cm}:name (n5 / name \\
\hspace{3cm}:op1 \cellcolor{rolecolor_2}{"Major"} \\
\hspace{3cm}:op2 \cellcolor{rolecolor_2}{"League"} \\
\hspace{3cm}:op3 \cellcolor{rolecolor_2}{"Soccer"})) \\
\hspace{2cm}:ARG1-of (c5 / cause-01 \\
\hspace{2.5cm}:ARG0 (h / have-org-role-91 \\
\hspace{3cm}:ARG0 f \\
\hspace{3cm}:ARG1 (o2 / organization \\
\hspace{3.5cm}:wiki \cellcolor{rolecolor_2}{"Western\_Conference"} \\
\hspace{3.5cm}:name (n6 / name \\
\hspace{4cm}:op1 \cellcolor{rolecolor_2}{"Western"} \\
\hspace{4cm}:op2 \cellcolor{rolecolor_2}{"Conference"})) \\
\hspace{3cm}:ARG2 (m2 / member))))) \\
\hline
\textbf{Reconstructed Context:} The Houston Dynamo is a professional soccer club based in Texas and competes in Major League Soccer. \\
\hline
\end{tabular}
\end{table}

\section{Prompts for Baselines}
\label{sec:prompt}

Drawing inspiration from the Alpaca\footnote{\url{https://crfm.stanford.edu/2023/03/13/alpaca.html}}, we set the prompt templates for baselines incorporating instructions of Keywords and Summary, which are presented in Prompt~\ref{prompt:baselines}.

\begin{tcolorbox}[
    colframe=black!75, 
    label=prompt:baselines, 
    fonttitle=\bfseries, 
    title={Prompt A1: The prompt for extracting the keywords and generating the summaries of the raw supporting documents as baselines.}, before upper=\refstepcounter{mytcolorbox}\label{prompt:baselines}
]
{[INST] <<SYS>> \\ Instruction (Keywords): \texttt{Extract a few keywords from the following content.}\\ 
Instruction (Summary): \texttt{Generate a short summary of the following content.}\\ 
\text{<</SYS>>}} \\
{Prompt = """Below is an instruction that describes a task, paired with an input that provides content.\\     \#\#\# Instruction: \{""" + \texttt{Instruction} + """\}\\    \#\#\# Input: \{""" + $\mathcal{D}$ + """\}\\     \#\#\# Response: """} \\
\text{[/INST]} 
\end{tcolorbox}

\section{Accuracy Details}
\label{sec:acc_details}

Table~\ref{tab:aca_PopQA} and Table~\ref{tab:aca_EQQA} showcase the comprehensive accuracy ($Acc$) results achieved by various context reconstruction methods on the PopQA and EntityQuestions datasets, respectively.

\begin{table}[htbp]\footnotesize
\centering
\caption{Accuracy ($Acc \uparrow$) on the PopQA dataset. The best results for each LLM with setting $K$ are in \textbf{bold}, and the next best results are in \uline{underlined}. $\Delta$ here represents the difference between the CoCR-RAG and Vanilla, and the \textcolor{applegreen}{increased} and \textcolor{downred}{decreased} $\Delta$ are marked differently. The best results for each of $K$ are \textcolor{bestcolor}{marked}.}
\label{tab:aca_PopQA}
\begin{adjustbox}{width=1\linewidth}
\begin{tabular}{c|c|*{10}{c}}
\hline
LLMs & \diagbox{$\mathcal{C}$}{$K$} & 1 & 2 & 3 & 4 & 5 & 6 & 7 & 8 & 9 & 10 \\
\hline
\multirow{7}{*}{\rotatebox{90}{G-1.3}} & Vanilla & 52.07 & 67.38 & 73.43 & 74.15 & 74.81 & 86.45 & \uline{93.06} & 89.19 & \uline{94.74} & 83.33 \\
 & Keywords & 23.71 & 29.93 & 44.66 & 51.50 & 55.43 & 70.97 & 61.11 & 75.68 & 68.42 & 83.33 \\
 & Summary & 27.89 & 39.46 & 50.81 & 56.84 & 64.73 & 67.10 & 72.22 & 72.97 & 78.95 & \textcolor{bestcolor}{\textbf{100.00}} \\
 & SelCon & 22.85 & 32.11 & 43.74 & 47.01 & 58.91 & 70.32 & 68.06 & 89.19 & 84.21 & 83.33 \\
 & LLMLingua & \textbf{56.99} & \uline{69.77} & \uline{77.03} & \textbf{78.63} & \uline{81.78} & \uline{89.68} & \textcolor{bestcolor}{\textbf{94.44}} & \textbf{91.89} & 94.73 & 83.33 \\
 & \cellcolor{markcolor} CoCR-RAG & \cellcolor{markcolor}\uline{56.81} & \cellcolor{markcolor}\textbf{70.02} & \cellcolor{markcolor}\textbf{77.38} & \cellcolor{markcolor}\uline{78.21} & \cellcolor{markcolor}\textbf{84.11} & \cellcolor{markcolor}\textbf{90.32} & \cellcolor{markcolor}\uline{93.06} & \cellcolor{markcolor}\textbf{91.89} & \cellcolor{markcolor}\textcolor{bestcolor}{\textbf{100.00}} & \cellcolor{markcolor}\textcolor{bestcolor}{\textbf{100.00}} \\
 & $\Delta$ & \textcolor{applegreen}{+4.74} & \textcolor{applegreen}{+2.64} & \textcolor{applegreen}{+3.95} & \textcolor{applegreen}{+4.06} & \textcolor{applegreen}{+9.30} & \textcolor{applegreen}{+3.87} & 0.00 & \textcolor{applegreen}{+2.70} & \textcolor{applegreen}{+5.26} & \textcolor{applegreen}{+16.67} \\
\hline
\multirow{7}{*}{\rotatebox{90}{G-2.7}} & Vanilla & \textbf{61.83} & \textbf{73.75} & 76.22 & 77.78 & 80.62 & \uline{87.10} & 84.72 & 89.19 & \textcolor{bestcolor}{\textbf{100.00}} & 83.33 \\
 & Keywords & 27.83 & 34.42 & 51.04 & 58.12 & 62.79 & 73.55 & 73.61 & 75.68 & 73.68 & \textcolor{bestcolor}{\textbf{100.00}} \\
 & Summary & 26.33 & 38.09 & 53.25 & 57.91 & 68.99 & 70.97 & 84.72 & 86.48 & 94.74 & 66.67 \\
 & SelCon & 36.12 & 45.53 & 53.36 & 60.90 & 69.38 & 74.19 & 80.56 & 91.89 & 94.74 & \textcolor{bestcolor}{\textbf{100.00}} \\
 & LLMLingua & \uline{56.01} & \uline{71.01} & \uline{77.26} & \uline{78.21} & \uline{81.78} & \uline{87.10} & \uline{91.67} & \uline{94.59} & \textcolor{bestcolor}{\textbf{100.00}} & \textcolor{bestcolor}{\textbf{100.00}} \\
 & \cellcolor{markcolor} CoCR-RAG & \cellcolor{markcolor}55.80 & \cellcolor{markcolor}69.05 & \cellcolor{markcolor}\textbf{77.84} & \cellcolor{markcolor}\textbf{79.06} & \cellcolor{markcolor}\textbf{86.82} & \cellcolor{markcolor}\textbf{90.32} & \cellcolor{markcolor}\textbf{93.06} & \cellcolor{markcolor}\textcolor{bestcolor}{\textbf{100.00}} & \cellcolor{markcolor}\textcolor{bestcolor}{\textbf{100.00}} & \cellcolor{markcolor}\textcolor{bestcolor}{\textbf{100.00}} \\
 & $\Delta$ & \textcolor{downred}{-6.03} & \textcolor{downred}{-4.70} & \textcolor{applegreen}{+1.62} & \textcolor{applegreen}{+1.28} & \textcolor{applegreen}{+6.20} & \textcolor{applegreen}{+3.22} & \textcolor{applegreen}{+8.34} & \textcolor{applegreen}{+10.81} & 0.00 & \textcolor{applegreen}{+16.67} \\
\hline
\multirow{7}{*}{\rotatebox{90}{G-j-6}} & Vanilla & \textbf{51.42} & 57.25 & 61.95 & 71.58 & \uline{75.58} & 80.00 & \uline{81.94} & 72.97 & \textbf{89.47} & \textcolor{bestcolor}{\textbf{100.00}} \\
 & Keywords & 20.91 & 22.02 & 27.96 & 33.76 & 30.23 & 34.84 & 37.50 & 37.84 & 36.84 & 50.00 \\
 & Summary & 21.38 & 29.67 & 40.26 & 38.03 & 44.96 & 48.39 & 56.94 & 54.05 & 73.68 & 83.33 \\
 & SelCon & 28.51 & 34.80 & 47.10 & 55.56 & 62.79 & 72.26 & 75.00 & \uline{78.38} & \textbf{89.47} & \textcolor{bestcolor}{\textbf{100.00}} \\
 & LLMLingua & \uline{49.00} & \uline{59.98} & \textbf{64.97} & \textbf{72.00} & \uline{75.58} & \uline{85.16} & 79.17 & \uline{78.38} & 84.21 & \textcolor{bestcolor}{\textbf{100.00}} \\
 & \cellcolor{markcolor} CoCR-RAG & \cellcolor{markcolor}48.97 & \cellcolor{markcolor}\textbf{61.35} & \cellcolor{markcolor}\uline{64.15} & \cellcolor{markcolor}\uline{71.79} & \cellcolor{markcolor}\textbf{76.74} & \cellcolor{markcolor}\textbf{85.81} & \cellcolor{markcolor}\textbf{83.33} & \cellcolor{markcolor}\textbf{94.59} & \cellcolor{markcolor}\textbf{89.47} & \cellcolor{markcolor}\textcolor{bestcolor}{\textbf{100.00}} \\
 & $\Delta$ & \textcolor{downred}{-2.45} & \textcolor{applegreen}{+4.10} & \textcolor{applegreen}{+2.20} & \textcolor{applegreen}{+0.21} & \textcolor{applegreen}{+1.16} & \textcolor{applegreen}{+5.81} & \textcolor{applegreen}{+1.39} & \textcolor{applegreen}{+21.62} & 0.00 & 0.00 \\
\hline
\multirow{7}{*}{\rotatebox{90}{O-1.3}} & Vanilla & \uline{66.60} & \textbf{76.14} & 78.54 & 80.34 & 78.68 & 85.81 & 91.67 & 89.19 & 94.74 & \textcolor{bestcolor}{\textbf{100.00}} \\
 & Keywords & 31.17 & 35.27 & 58.00 & 65.17 & 72.48 & 81.94 & 87.50 & 81.08 & 89.47 & \textcolor{bestcolor}{\textbf{100.00}} \\
 & Summary & 31.05 & 45.49 & 63.34 & 66.24 & 79.07 & 81.94 & 90.28 & 89.19 & \textcolor{bestcolor}{\textbf{100.00}} & \textcolor{bestcolor}{\textbf{100.00}} \\
 & SelCon & 45.03 & 55.79 & 64.85 & 72.44 & 78.68 & 79.35 & 90.28 & \textbf{94.59} & 94.74 & \textcolor{bestcolor}{\textbf{100.00}} \\
 & LLMLingua & \textbf{67.94} & 68.96 & \textbf{80.74} & \uline{83.55} & \uline{85.27} & \uline{90.97} & \textcolor{bestcolor}{\textbf{94.44}} & \uline{91.89} & 94.74 & \textcolor{bestcolor}{\textbf{100.00}} \\
 & \cellcolor{markcolor} CoCR-RAG & \cellcolor{markcolor}63.47 & \cellcolor{markcolor}\uline{73.45} & \cellcolor{markcolor}\uline{78.89} & \cellcolor{markcolor}\textbf{85.68} & \cellcolor{markcolor}\textbf{86.82} & \cellcolor{markcolor}\textbf{92.26} & \cellcolor{markcolor}\uline{93.06} & \cellcolor{markcolor}\uline{91.89} & \cellcolor{markcolor}\textcolor{bestcolor}{\textbf{100.00}} & \cellcolor{markcolor}\textcolor{bestcolor}{\textbf{100.00}} \\
 & $\Delta$ & \textcolor{downred}{-3.13} & \textcolor{downred}{-2.69} & \textcolor{applegreen}{+0.35} & \textcolor{applegreen}{+5.34} & \textcolor{applegreen}{+8.14} & \textcolor{applegreen}{+6.45} & \textcolor{applegreen}{+1.39} & \textcolor{applegreen}{+2.70} & \textcolor{applegreen}{+5.26} & 0.00 \\
\hline
\multirow{7}{*}{\rotatebox{90}{O-2.7}} & Vanilla & \textbf{67.49} & \uline{77.64} & 79.70 & 79.91 & 84.88 & 88.39 & \uline{91.67} & \uline{94.59} & 94.74 & \textcolor{bestcolor}{\textbf{100.00}} \\
 & Keywords & 31.35 & 35.49 & 54.29 & 61.32 & 66.67 & 82.58 & 83.33 & 78.38 & 89.47 & \textcolor{bestcolor}{\textbf{100.00}} \\
 & Summary & 31.37 & 43.65 & 63.57 & 67.09 & 75.19 & 72.26 & 81.94 & 67.57 & 89.47 & 66.67 \\
 & SelCon & 34.12 & 45.53 & 54.99 & 62.39 & 72.09 & 82.58 & 84.72 & 91.89 & 89.47 & \textcolor{bestcolor}{\textbf{100.00}} \\
 & LLMLingua & 62.99 & \textcolor{bestcolor}{\textbf{78.79}} & \textcolor{bestcolor}{\textbf{83.18}} & \uline{82.48} & \uline{86.43} & \uline{89.03} & 90.28 & 91.89 & \textcolor{bestcolor}{\textbf{100.00}} & \textcolor{bestcolor}{\textbf{100.00}} \\
 & \cellcolor{markcolor} CoCR-RAG & \cellcolor{markcolor}\uline{65.40} & \cellcolor{markcolor}75.07 & \cellcolor{markcolor}\uline{81.79} & \cellcolor{markcolor}\textbf{85.26} & \cellcolor{markcolor}\textbf{90.31} & \cellcolor{markcolor}\textcolor{bestcolor}{\textbf{96.13}} & \cellcolor{markcolor}\textbf{93.06} & \cellcolor{markcolor}\textbf{97.30} & \cellcolor{markcolor}\textcolor{bestcolor}{\textbf{100.00}} & \cellcolor{markcolor}\textcolor{bestcolor}{\textbf{100.00}} \\
 & $\Delta$ & \textcolor{downred}{-2.09} & \textcolor{downred}{-2.57} & \textcolor{applegreen}{+2.09} & \textcolor{applegreen}{+5.35} & \textcolor{applegreen}{+5.43} & \textcolor{applegreen}{+7.74} & \textcolor{applegreen}{+1.39} & \textcolor{applegreen}{+2.71} & \textcolor{applegreen}{+5.26} & 0.00 \\
\hline
\multirow{7}{*}{\rotatebox{90}{O-6.7}} & Vanilla & 35.79 & 69.26 & \uline{75.17} & \uline{80.56} & 83.33 & 87.10 & 90.28 & \uline{89.19} & 94.74 & \textcolor{bestcolor}{\textbf{100.00}} \\
 & Keywords & 33.70 & 40.36 & 56.96 & 63.89 & 67.83 & 74.84 & 79.17 & 81.08 & 89.47 & \textcolor{bestcolor}{\textbf{100.00}} \\
 & Summary & 28.48 & 37.15 & 47.79 & 44.87 & 53.49 & 58.06 & 54.17 & 56.76 & 57.89 & 83.33 \\
 & SelCon & 49.24 & 64.39 & 70.77 & 76.71 & 81.40 & 87.10 & \uline{91.67} & 86.49 & \textcolor{bestcolor}{\textbf{100.00}} & 83.33 \\
 & LLMLingua & \uline{56.99} & \uline{69.39} & 74.01 & 79.49 & \uline{84.11} & \uline{90.32} & 90.28 & 86.49 & \textcolor{bestcolor}{\textbf{100.00}} & \textcolor{bestcolor}{\textbf{100.00}} \\
 & \cellcolor{markcolor} CoCR-RAG & \cellcolor{markcolor}\textbf{60.60} & \cellcolor{markcolor}\textbf{71.14} & \cellcolor{markcolor}\textbf{79.47} & \cellcolor{markcolor}\textbf{82.05} & \cellcolor{markcolor}\textbf{85.27} & \cellcolor{markcolor}\textbf{91.61} & \cellcolor{markcolor}\textcolor{bestcolor}{\textbf{94.44}} & \cellcolor{markcolor}\textcolor{bestcolor}{\textbf{100.00}} & \cellcolor{markcolor}\textcolor{bestcolor}{\textbf{100.00}} & \cellcolor{markcolor}\textcolor{bestcolor}{\textbf{100.00}} \\
 & $\Delta$ & \textcolor{applegreen}{+24.81} & \textcolor{applegreen}{+1.88} & \textcolor{applegreen}{+4.30} & \textcolor{applegreen}{+1.49} & \textcolor{applegreen}{+1.94} & \textcolor{applegreen}{+4.51} & \textcolor{applegreen}{+4.16} & \textcolor{applegreen}{+10.81} & \textcolor{applegreen}{+5.26} & 0.00 \\
\hline
\multirow{7}{*}{\rotatebox{90}{b-560}} & Vanilla & \textbf{56.25} & \textbf{63.57} & 64.04 & 72.86 & 72.87 & \uline{80.00} & 79.17 & 81.08 & 89.47 & \textcolor{bestcolor}{\textbf{100.00}} \\
 & Keywords & 24.04 & 30.65 & 41.76 & 48.72 & 50.39 & 60.65 & 62.50 & 70.27 & 73.68 & 83.33 \\
 & Summary & 26.45 & 38.44 & 55.57 & 63.03 & 72.09 & 74.84 & 80.56 & \uline{83.78} & \textbf{94.74} & \textcolor{bestcolor}{\textbf{100.00}} \\
 & SelCon & 23.56 & 24.03 & 31.79 & 36.97 & 46.51 & 55.48 & 56.94 & 67.57 & 78.95 & 66.67 \\
 & LLMLingua & 48.46 & \uline{63.23} & \uline{65.78} & \uline{73.50} & \uline{73.64} & 79.35 & \uline{81.94} & \uline{83.78} & 89.47 & \textcolor{bestcolor}{\textbf{100.00}} \\
 & \cellcolor{markcolor} CoCR-RAG & \cellcolor{markcolor}\uline{49.42} & \cellcolor{markcolor}61.27 & \cellcolor{markcolor}\textbf{68.56} & \cellcolor{markcolor}\textbf{74.57} & \cellcolor{markcolor}\textbf{80.62} & \cellcolor{markcolor}\textbf{82.58} & \cellcolor{markcolor}\textbf{84.72} & \cellcolor{markcolor}\textbf{94.59} & \cellcolor{markcolor}\textbf{94.74} & \cellcolor{markcolor}\textcolor{bestcolor}{\textbf{100.00}} \\
 & $\Delta$ & \textcolor{downred}{-6.83} & \textcolor{downred}{-2.30} & \textcolor{applegreen}{+4.52} & \textcolor{applegreen}{+1.71} & \textcolor{applegreen}{+7.75} & \textcolor{applegreen}{+2.58} & \textcolor{applegreen}{+5.55} & \textcolor{applegreen}{+13.51} & \textcolor{applegreen}{+5.27} & 0.00 \\
\hline
\multirow{7}{*}{\rotatebox{90}{b-1b1}} & Vanilla & \textbf{64.84} & \textbf{72.85} & \uline{77.96} & 78.63 & 75.19 & 87.10 & 88.89 & \uline{94.59} & \textbf{94.74} & 83.33 \\
 & Keywords & 30.18 & 41.98 & 57.89 & 64.32 & 70.16 & 80.65 & 79.17 & 89.19 & 78.95 & \textcolor{bestcolor}{\textbf{100.00}} \\
 & Summary & 29.44 & 44.46 & 61.60 & 62.61 & 71.71 & 78.71 & \uline{90.28} & 81.08 & 78.95 & \textcolor{bestcolor}{\textbf{100.00}} \\
 & SelCon & 45.87 & 55.96 & 61.48 & 67.31 & 73.64 & 80.65 & 83.33 & 91.89 & 84.21 & \textcolor{bestcolor}{\textbf{100.00}} \\
 & LLMLingua & 59.94 & 71.78 & 75.63 & \textbf{80.98} & \uline{82.95} & \textbf{89.68} & \uline{90.28} & \uline{94.59} & \textbf{94.74} & \textcolor{bestcolor}{\textbf{100.00}} \\
 & \cellcolor{markcolor} CoCR-RAG & \cellcolor{markcolor}\uline{60.04} & \cellcolor{markcolor}\uline{72.55} & \cellcolor{markcolor}\textbf{78.07} & \cellcolor{markcolor}\uline{79.91} & \cellcolor{markcolor}\textbf{85.27} & \cellcolor{markcolor}\uline{89.03} & \cellcolor{markcolor}\textbf{93.06} & \cellcolor{markcolor}\textbf{97.30} & \cellcolor{markcolor}\textbf{94.74} & \cellcolor{markcolor}\textcolor{bestcolor}{\textbf{100.00}} \\
 & $\Delta$ & \textcolor{downred}{-4.80} & \textcolor{downred}{-0.30} & \textcolor{applegreen}{+0.11} & \textcolor{applegreen}{+1.28} & \textcolor{applegreen}{+10.08} & \textcolor{applegreen}{+1.93} & \textcolor{applegreen}{+4.17} & \textcolor{applegreen}{+2.71} & 0.00 & \textcolor{applegreen}{+16.67} \\
\hline
\multirow{7}{*}{\rotatebox{90}{b-1b7}} & Vanilla & \textbf{66.87} & \textbf{74.90} & \uline{75.17} & 77.78 & \uline{79.84} & 86.45 & 84.72 & 89.19 & 89.47 & \textcolor{bestcolor}{\textbf{100.00}} \\
 & Keywords & 31.85 & 41.56 & 56.03 & 69.87 & 70.16 & 83.87 & 79.17 & 89.19 & 89.47 & \textcolor{bestcolor}{\textbf{100.00}} \\
 & Summary & 30.72 & 43.91 & 61.60 & 68.16 & 78.29 & 74.84 & \textbf{87.50} & \textbf{97.30} & \textbf{94.74} & \textcolor{bestcolor}{\textbf{100.00}} \\
 & SelCon & 41.87 & 49.51 & 55.68 & 63.46 & 70.54 & 84.52 & 77.78 & 86.49 & 84.21 & \textcolor{bestcolor}{\textbf{100.00}} \\
 & LLMLingua & 61.02 & \uline{73.62} & \uline{75.17} & \uline{78.21} & 76.74 & \uline{87.74} & \textbf{87.50} & \uline{91.89} & 89.47 & \textcolor{bestcolor}{\textbf{100.00}} \\
 & \cellcolor{markcolor} CoCR-RAG & \cellcolor{markcolor}\uline{62.00} & \cellcolor{markcolor}72.98 & \cellcolor{markcolor}\textbf{77.84} & \cellcolor{markcolor}\textbf{78.85} & \cellcolor{markcolor}\textbf{89.15} & \cellcolor{markcolor}\textbf{90.97} & \cellcolor{markcolor}86.11 & \cellcolor{markcolor}\uline{91.89} & \cellcolor{markcolor}\textbf{94.74} & \cellcolor{markcolor}\textcolor{bestcolor}{\textbf{100.00}} \\
 & $\Delta$ & \textcolor{downred}{-4.87} & \textcolor{downred}{-1.92} & \textcolor{applegreen}{+2.67} & \textcolor{applegreen}{+1.07} & \textcolor{applegreen}{+9.31} & \textcolor{applegreen}{+4.52} & \textcolor{applegreen}{+1.39} & \textcolor{applegreen}{+2.70} & \textcolor{applegreen}{+5.27} & 0.00 \\
\hline
\multirow{7}{*}{\rotatebox{90}{b-3}} & Vanilla & \textbf{67.97} & \textbf{74.35} & \uline{77.38} & \uline{75.00} & \uline{82.95} & \uline{88.39} & \uline{88.89} & 89.19 & 89.47 & \textcolor{bestcolor}{\textbf{100.00}} \\
 & Keywords & 33.64 & 41.30 & 59.16 & 66.03 & 70.54 & 83.23 & 75.00 & 86.49 & \textbf{94.74} & \textcolor{bestcolor}{\textbf{100.00}} \\
 & Summary & 30.39 & 42.92 & 60.09 & 67.09 & 74.81 & 78.06 & 83.33 & 81.08 & 89.47 & \textcolor{bestcolor}{\textbf{100.00}} \\
 & SelCon & 49.09 & 55.24 & 66.24 & 69.87 & 72.87 & 83.87 & 84.72 & \textbf{94.59} & 89.47 & \textcolor{bestcolor}{\textbf{100.00}} \\
 & LLMLingua & \uline{65.23} & \uline{72.89} & 74.83 & 74.36 & 77.91 & 85.16 & 87.50 & 91.89 & \textbf{94.74} & \textcolor{bestcolor}{\textbf{100.00}} \\
 & \cellcolor{markcolor} CoCR-RAG & \cellcolor{markcolor}63.02 & \cellcolor{markcolor}72.47 & \cellcolor{markcolor}\textbf{78.31} & \cellcolor{markcolor}\textbf{80.34} & \cellcolor{markcolor}\textbf{88.76} & \cellcolor{markcolor}\textbf{89.68} & \cellcolor{markcolor}\textcolor{bestcolor}{\textbf{94.44}} & \cellcolor{markcolor}\textbf{94.59} & \cellcolor{markcolor}\textbf{94.74} & \cellcolor{markcolor}\textcolor{bestcolor}{\textbf{100.00}} \\
 & $\Delta$ & \textcolor{downred}{-4.95} & \textcolor{downred}{-1.88} & \textcolor{applegreen}{+0.93} & \textcolor{applegreen}{+5.34} & \textcolor{applegreen}{+5.81} & \textcolor{applegreen}{+1.29} & \textcolor{applegreen}{+5.55} & \textcolor{applegreen}{+5.40} & \textcolor{applegreen}{+5.27} & 0.00 \\
\hline
\multirow{7}{*}{\rotatebox{90}{L-7}} & Vanilla & 58.30 & 64.81 & 63.81 & 74.36 & 77.91 & 80.65 & 80.56 & 89.19 & \uline{89.47} & \textcolor{bestcolor}{\textbf{100.00}} \\
 & Keywords & 45.84 & 57.63 & 68.33 & 75.00 & 78.29 & 77.42 & 84.72 & 86.49 & \uline{89.47} & \textcolor{bestcolor}{\textbf{100.00}} \\
 & Summary & 38.09 & 44.63 & 57.08 & 58.76 & 62.79 & 63.23 & 66.67 & 64.86 & 73.68 & 66.67 \\
 & SelCon & 62.75 & 73.54 & \uline{79.81} & 83.12 & 85.27 & \uline{88.39} & \uline{91.67} & 91.89 & \uline{89.47} & \textcolor{bestcolor}{\textbf{100.00}} \\
 & LLMLingua & \textbf{68.00} & \uline{74.99} & 78.89 & \uline{85.68} & \uline{86.43} & \textbf{89.03} & 88.89 & \textbf{94.59} & \uline{89.47} & \textcolor{bestcolor}{\textbf{100.00}} \\
 & \cellcolor{markcolor} CoCR-RAG & \cellcolor{markcolor}\uline{67.13} & \cellcolor{markcolor}\textbf{76.91} & \cellcolor{markcolor}\textbf{80.97} & \cellcolor{markcolor}\textcolor{bestcolor}{\textbf{86.32}} & \cellcolor{markcolor}\textbf{88.37} & \cellcolor{markcolor}\uline{88.39} & \cellcolor{markcolor}\textbf{93.06} & \cellcolor{markcolor}\textbf{94.59} & \cellcolor{markcolor}\textbf{94.74} & \cellcolor{markcolor}\textcolor{bestcolor}{\textbf{100.00}} \\
 & $\Delta$ & \textcolor{applegreen}{+8.83} & \textcolor{applegreen}{+12.10} & \textcolor{applegreen}{+17.16} & \textcolor{applegreen}{+11.96} & \textcolor{applegreen}{+10.46} & \textcolor{applegreen}{+7.74} & \textcolor{applegreen}{+12.50} & \textcolor{applegreen}{+5.40} & \textcolor{applegreen}{+5.27} & 0.00 \\
\hline
\multirow{7}{*}{\rotatebox{90}{L-13}} & Vanilla & 60.87 & \uline{77.13} & 77.39 & 78.42 & 79.46 & 83.87 & 81.94 & 91.89 & \uline{89.47} & \textcolor{bestcolor}{\textbf{100.00}} \\
 & Keywords & 47.45 & 58.87 & 71.81 & 79.27 & 82.95 & 84.52 & 87.50 & 89.19 & \textbf{94.74} & \textcolor{bestcolor}{\textbf{100.00}} \\
 & Summary & 41.49 & 51.52 & 58.12 & 59.83 & 63.95 & 69.03 & 70.83 & 89.19 & \uline{89.47} & 83.33 \\
 & SelCon & 69.10 & 76.06 & 80.74 & \textbf{83.76} & 85.66 & 88.39 & \textbf{90.28} & 94.59 & \uline{89.47} & \textcolor{bestcolor}{\textbf{100.00}} \\
 & LLMLingua & \textcolor{bestcolor}{\textbf{70.00}} & 72.98 & \uline{81.09} & 80.98 & \uline{87.98} & \textbf{89.68} & 88.89 & \textbf{97.30} & \uline{89.47} & \textcolor{bestcolor}{\textbf{100.00}} \\
 & \cellcolor{markcolor} CoCR-RAG & \cellcolor{markcolor}\uline{69.49} & \cellcolor{markcolor}\textbf{77.30} & \cellcolor{markcolor}\textbf{81.79} & \cellcolor{markcolor}\uline{82.69} & \cellcolor{markcolor}\textcolor{bestcolor}{\textbf{90.70}} & \cellcolor{markcolor}\textbf{89.68} & \cellcolor{markcolor}\textbf{90.28} & \cellcolor{markcolor}\textbf{97.30} & \cellcolor{markcolor}\uline{89.47} & \cellcolor{markcolor}\textcolor{bestcolor}{\textbf{100.00}} \\
 & $\Delta$ & \textcolor{applegreen}{+8.62} & \textcolor{applegreen}{+0.17} & \textcolor{applegreen}{+4.40} & \textcolor{applegreen}{+4.27} & \textcolor{applegreen}{+11.24} & \textcolor{applegreen}{+5.81} & \textcolor{applegreen}{+8.34} & \textcolor{applegreen}{+5.41} & 0.00 & 0.00 \\
\hline

\end{tabular}
\end{adjustbox}
\end{table}

\begin{table}[!htbp]\footnotesize
\centering
\caption{Accuracy ($Acc \uparrow$) the EntityQuestions dataset. The definitions of symbols are the same with the Table~\ref{tab:aca_PopQA}.}
\label{tab:aca_EQQA}
\begin{adjustbox}{width=1\linewidth}
\begin{tabular}{c|c|*{10}{c}}
\hline
LLMs & \diagbox{$\mathcal{C}$}{$K$} & 1 & 2 & 3 & 4 & 5 & 6 & 7 & 8 & 9 & 10 \\
\hline
\multirow{7}{*}{\rotatebox{90}{G-1.3}} & Vanilla & 44.22 & 52.77 & 59.49 & 59.71 & 63.34 & \uline{64.73} & 61.35 & \textbf{72.53} & 72.06 & \uline{75.00} \\
 & Keywords & 19.48 & 27.29 & 29.27 & 36.45 & 34.25 & 40.63 & 35.58 & 20.88 & 36.76 & 41.67 \\
 & Summary & 27.75 & 34.54 & 39.56 & 42.68 & 46.58 & 44.20 & 42.94 & 37.36 & 45.59 & 45.83 \\
 & SelCon & 21.74 & 30.06 & 28.01 & 32.37 & 37.33 & 39.29 & 38.66 & 36.26 & 45.59 & 37.50 \\
 & LLMLingua & \uline{45.02} & \uline{59.26} & \textbf{63.77} & \uline{63.55} & \uline{63.36} & 64.29 & \uline{63.80} & 63.74 & \textbf{77.94} & \uline{75.00} \\
 & \cellcolor{markcolor} CoCR-RAG & \cellcolor{markcolor}\textbf{47.57} & \cellcolor{markcolor}\textbf{62.02} & \cellcolor{markcolor}\uline{62.34} & \cellcolor{markcolor}\textbf{66.91} & \cellcolor{markcolor}\textbf{73.63} & \cellcolor{markcolor}\textbf{71.43} & \cellcolor{markcolor}\textbf{75.46} & \cellcolor{markcolor}\uline{70.33} & \cellcolor{markcolor}\uline{76.47} & \cellcolor{markcolor}\textbf{83.33} \\
 & $\Delta$ & \textcolor{applegreen}{+3.35} & \textcolor{applegreen}{+9.25} & \textcolor{applegreen}{+2.85} & \textcolor{applegreen}{+7.20} & \textcolor{applegreen}{+10.29} & \textcolor{applegreen}{+6.70} & \textcolor{applegreen}{+14.11} & \textcolor{downred}{-2.20} & \textcolor{applegreen}{+4.41} & \textcolor{applegreen}{+8.33} \\
\hline
\multirow{7}{*}{\rotatebox{90}{G-2.7}} & Vanilla & \uline{52.04} & 59.83 & \textbf{66.77} & \uline{69.30} & \uline{72.95} & 73.66 & \uline{71.78} & \uline{74.73} & \uline{77.94} & 75.00 \\
 & Keywords & 27.07 & 34.54 & 38.45 & 45.80 & 41.44 & 51.34 & 43.56 & 36.26 & 38.23 & 45.83 \\
 & Summary & 26.73 & 36.35 & 40.98 & 46.28 & 47.60 & 50.00 & 51.53 & 48.35 & 55.88 & 29.17 \\
 & SelCon & 31.65 & 39.22 & 43.51 & 47.24 & 50.34 & 50.89 & 48.47 & 38.46 & 52.94 & 45.83 \\
 & LLMLingua & \textbf{52.10} & \uline{62.11} & 62.66 & 67.39 & 68.49 & \uline{75.89} & 65.64 & \uline{74.73} & \uline{77.94} & \textbf{79.17} \\
 & \cellcolor{markcolor} CoCR-RAG & \cellcolor{markcolor}49.38 & \cellcolor{markcolor}\textbf{62.40} & \cellcolor{markcolor}\uline{64.40} & \cellcolor{markcolor}\textbf{71.46} & \cellcolor{markcolor}\textbf{76.37} & \cellcolor{markcolor}\textbf{78.13} & \cellcolor{markcolor}\textbf{77.30} & \cellcolor{markcolor}\textbf{75.82} & \cellcolor{markcolor}\textbf{82.35} & \cellcolor{markcolor}\textbf{79.17} \\
 & $\Delta$ & \textcolor{downred}{-2.66} & \textcolor{applegreen}{+2.57} & \textcolor{downred}{-2.37} & \textcolor{applegreen}{+2.16} & \textcolor{applegreen}{+3.42} & \textcolor{applegreen}{+4.47} & \textcolor{applegreen}{+5.52} & \textcolor{applegreen}{+1.09} & \textcolor{applegreen}{+4.41} & \textcolor{applegreen}{+4.17} \\
\hline
\multirow{7}{*}{\rotatebox{90}{G-j-6}} & Vanilla & \textbf{44.96} & 52.58 & 58.39 & \uline{62.59} & 68.15 & 70.09 & 73.62 & 73.63 & \textbf{79.41} & \uline{83.33} \\
 & Keywords & 18.12 & 19.47 & 18.20 & 21.82 & 20.89 & 22.32 & 23.31 & 25.27 & 22.06 & 29.17 \\
 & Summary & 23.95 & 26.62 & 26.11 & 28.06 & 30.48 & 25.89 & 30.67 & 25.27 & 23.53 & 37.50 \\
 & SelCon & 26.73 & 35.97 & 37.34 & 39.57 & 47.60 & 51.79 & 52.15 & 50.55 & 48.53 & 50.00 \\
 & LLMLingua & 40.32 & \uline{53.53} & \textbf{62.50} & 61.60 & \textbf{72.60} & \uline{71.88} & \uline{74.23} & \uline{74.73} & \uline{77.94} & 75.00 \\
 & \cellcolor{markcolor} CoCR-RAG & \cellcolor{markcolor}\uline{40.71} & \cellcolor{markcolor}\textbf{54.20} & \cellcolor{markcolor}\uline{61.55} & \cellcolor{markcolor}\textbf{65.47} & \cellcolor{markcolor}\uline{70.89} & \cellcolor{markcolor}\textbf{72.32} & \cellcolor{markcolor}\textbf{80.37} & \cellcolor{markcolor}\textbf{75.82} & \cellcolor{markcolor}\uline{77.94} & \cellcolor{markcolor}\textbf{87.50} \\
 & $\Delta$ & \textcolor{downred}{-4.25} & \textcolor{applegreen}{+1.62} & \textcolor{applegreen}{+3.16} & \textcolor{applegreen}{+2.88} & \textcolor{applegreen}{+2.74} & \textcolor{applegreen}{+2.23} & \textcolor{applegreen}{+6.75} & \textcolor{applegreen}{+2.19} & \textcolor{downred}{-1.47} & \textcolor{applegreen}{+4.17} \\
\hline
\multirow{7}{*}{\rotatebox{90}{O-1.3}} & Vanilla & \textbf{56.40} & 64.31 & 65.98 & 67.15 & 70.89 & 70.54 & 68.10 & 69.23 & \textbf{73.53} & 75.00 \\
 & Keywords & 30.18 & 31.97 & 42.56 & 46.76 & 57.53 & 51.79 & 46.01 & 49.45 & 50.00 & 58.33 \\
 & Summary & 32.50 & 43.99 & 50.95 & 55.64 & 65.41 & 64.73 & 60.74 & 62.64 & 64.71 & 58.33 \\
 & SelCon & 37.94 & 44.18 & 45.89 & 51.32 & 60.96 & 59.38 & 56.44 & 58.24 & 63.24 & 66.67 \\
 & LLMLingua & \uline{56.29} & \uline{64.89} & \textbf{68.04} & \uline{69.06} & \uline{71.92} & \uline{73.21} & \uline{71.78} & \uline{72.53} & \textbf{73.53} & \textbf{83.33} \\
 & \cellcolor{markcolor} CoCR-RAG & \cellcolor{markcolor}56.12 & \cellcolor{markcolor}\textbf{65.36} & \cellcolor{markcolor}\uline{67.09} & \cellcolor{markcolor}\textbf{72.42} & \cellcolor{markcolor}\textbf{75.34} & \cellcolor{markcolor}\textbf{78.57} & \cellcolor{markcolor}\textbf{75.46} & \cellcolor{markcolor}\textbf{78.02} & \cellcolor{markcolor}72.06 & \cellcolor{markcolor}\textbf{83.33} \\
 & $\Delta$ & \textcolor{downred}{-0.28} & \textcolor{applegreen}{+1.05} & \textcolor{applegreen}{+1.11} & \textcolor{applegreen}{+5.27} & \textcolor{applegreen}{+4.45} & \textcolor{applegreen}{+8.03} & \textcolor{applegreen}{+7.36} & \textcolor{applegreen}{+8.79} & \textcolor{downred}{-1.47} & \textcolor{applegreen}{+8.33} \\
\hline
\multirow{7}{*}{\rotatebox{90}{O-2.7}} & Vanilla & \textbf{59.51} & 66.89 & 70.73 & 68.35 & 76.37 & 74.55 & 77.91 & \uline{75.82} & 79.41 & 83.33 \\
 & Keywords & 33.86 & 36.64 & 42.56 & 51.32 & 54.11 & 55.36 & 58.28 & 48.35 & 50.00 & 54.17 \\
 & Summary & 32.21 & 45.90 & 49.53 & 52.04 & 56.51 & 56.25 & 51.53 & 41.76 & 52.94 & 54.17 \\
 & SelCon & 37.26 & 44.27 & 44.62 & 53.24 & 57.53 & 61.16 & 58.90 & 57.14 & 51.47 & 79.17 \\
 & LLMLingua & 58.21 & \uline{69.75} & \textbf{74.05} & \uline{71.70} & \uline{77.05} & \uline{76.79} & \textcolor{bestcolor}{\textbf{82.82}} & 74.73 & \textbf{83.82} & \uline{87.50} \\
 & \cellcolor{markcolor} CoCR-RAG & \cellcolor{markcolor}\uline{58.38} & \cellcolor{markcolor}\textbf{71.37} & \cellcolor{markcolor}\uline{72.15} & \cellcolor{markcolor}\textbf{76.74} & \cellcolor{markcolor}\textcolor{bestcolor}{\textbf{81.51}} & \cellcolor{markcolor}\textcolor{bestcolor}{\textbf{82.59}} & \cellcolor{markcolor}\textcolor{bestcolor}{\textbf{82.82}} & \cellcolor{markcolor}\textbf{79.12} & \cellcolor{markcolor}\textbf{83.82} & \cellcolor{markcolor}\textcolor{bestcolor}{\textbf{91.67}} \\
 & $\Delta$ & \textcolor{downred}{-1.13} & \textcolor{applegreen}{+4.48} & \textcolor{applegreen}{+1.42} & \textcolor{applegreen}{+8.39} & \textcolor{applegreen}{+5.14} & \textcolor{applegreen}{+8.04} & \textcolor{applegreen}{+4.91} & \textcolor{applegreen}{+3.30} & \textcolor{applegreen}{+4.41} & \textcolor{applegreen}{+8.34} \\
\hline
\multirow{7}{*}{\rotatebox{90}{O-6.7}} & Vanilla & 33.18 & 56.39 & 70.09 & 69.78 & 75.34 & \uline{78.13} & 74.85 & 80.22 & \uline{82.36} & \uline{79.17} \\
 & Keywords & 32.33 & 35.30 & 43.83 & 48.20 & 49.66 & 55.36 & 57.06 & 54.95 & 48.53 & 62.50 \\
 & Summary & 30.63 & 36.35 & 37.34 & 35.25 & 39.04 & 35.71 & 33.13 & 36.26 & 41.18 & 37.50 \\
 & SelCon & 47.90 & 56.87 & 62.66 & 64.51 & 66.44 & 75.89 & 73.62 & 72.53 & 80.88 & 70.83 \\
 & LLMLingua & \textbf{58.78} & \textbf{70.52} & \textbf{74.53} & \textbf{75.78} & \uline{76.37} & 77.68 & \uline{77.30} & \uline{81.32} & 82.35 & \textbf{83.33} \\
 & \cellcolor{markcolor} CoCR-RAG & \cellcolor{markcolor}\uline{54.59} & \cellcolor{markcolor}\uline{67.94} & \cellcolor{markcolor}\uline{72.78} & \cellcolor{markcolor}\textbf{75.78} & \cellcolor{markcolor}\textbf{78.77} & \cellcolor{markcolor}\textbf{79.02} & \cellcolor{markcolor}\textbf{80.98} & \cellcolor{markcolor}\textbf{82.42} & \cellcolor{markcolor}\textbf{83.82} & \cellcolor{markcolor}\uline{79.17} \\
 & $\Delta$ & \textcolor{applegreen}{+21.41} & \textcolor{applegreen}{+11.55} & \textcolor{applegreen}{+2.69} & \textcolor{applegreen}{+6.00} & \textcolor{applegreen}{+3.43} & \textcolor{applegreen}{+0.89} & \textcolor{applegreen}{+6.13} & \textcolor{applegreen}{+2.20} & \textcolor{applegreen}{+1.46} & 0.00 \\
\hline
\multirow{7}{*}{\rotatebox{90}{b-560}} & Vanilla & \uline{41.39} & 51.15 & 52.53 & \uline{58.99} & \uline{60.96} & \textbf{65.18} & \uline{58.28} & 54.95 & 60.29 & 62.50 \\
 & Keywords & 21.57 & 29.58 & 32.12 & 33.81 & 37.33 & 41.07 & 38.04 & 30.77 & 35.29 & 45.83 \\
 & Summary & 25.71 & 35.69 & 41.14 & 45.32 & 48.63 & 50.00 & 44.17 & 49.45 & 36.76 & 50.00 \\
 & SelCon & 21.40 & 24.14 & 18.67 & 21.10 & 29.79 & 30.35 & 30.06 & 29.67 & 29.41 & 33.33 \\
 & LLMLingua & 40.93 & \uline{51.81} & \uline{54.11} & 58.75 & 58.90 & 62.50 & 56.44 & \uline{67.03} & \uline{70.59} & \uline{66.67} \\
 & \cellcolor{markcolor} CoCR-RAG & \cellcolor{markcolor}\textbf{41.73} & \cellcolor{markcolor}\textbf{55.73} & \cellcolor{markcolor}\textbf{58.39} & \cellcolor{markcolor}\textbf{60.19} & \cellcolor{markcolor}\textbf{63.01} & \cellcolor{markcolor}\uline{62.95} & \cellcolor{markcolor}\textbf{63.80} & \cellcolor{markcolor}\textbf{75.82} & \cellcolor{markcolor}\textbf{75.00} & \cellcolor{markcolor}\textbf{79.17} \\
 & $\Delta$ & \textcolor{applegreen}{+0.34} & \textcolor{applegreen}{+4.58} & \textcolor{applegreen}{+5.86} & \textcolor{applegreen}{+1.20} & \textcolor{applegreen}{+2.05} & \textcolor{downred}{-2.23} & \textcolor{applegreen}{+5.52} & \textcolor{applegreen}{+20.87} & \textcolor{applegreen}{+14.71} & \textcolor{applegreen}{+16.67} \\
\hline
\multirow{7}{*}{\rotatebox{90}{b-1b1}} & Vanilla & \uline{53.06} & \uline{60.69} & \textbf{68.35} & \textbf{71.70} & \uline{72.26} & 70.98 & 67.48 & 70.33 & 73.53 & \textbf{83.33} \\
 & Keywords & 26.50 & 38.65 & 43.83 & 51.80 & 50.34 & 58.93 & 51.53 & 42.86 & 52.94 & 75.00 \\
 & Summary & 31.77 & 40.84 & 44.46 & 47.00 & 52.40 & 50.45 & 45.40 & 49.45 & 55.88 & 37.50 \\
 & SelCon & 37.88 & 44.75 & 48.73 & 50.36 & 56.51 & 60.27 & 56.44 & 47.25 & 57.35 & 66.67 \\
 & LLMLingua & 51.81 & 59.06 & \uline{66.93} & 66.91 & 71.23 & \uline{72.77} & \uline{69.33} & \uline{71.43} & \textbf{75.00} & \textbf{83.33} \\
 & \cellcolor{markcolor} CoCR-RAG & \cellcolor{markcolor}\textbf{54.13} & \cellcolor{markcolor}\textbf{63.45} & \cellcolor{markcolor}\uline{66.93} & \cellcolor{markcolor}\uline{69.30} & \cellcolor{markcolor}\textbf{76.03} & \cellcolor{markcolor}\textbf{78.57} & \cellcolor{markcolor}\textbf{75.46} & \cellcolor{markcolor}\textbf{74.73} & \cellcolor{markcolor}\textbf{75.00} & \cellcolor{markcolor}\textbf{83.33} \\
 & $\Delta$ & \textcolor{applegreen}{+1.07} & \textcolor{applegreen}{+2.76} & \textcolor{downred}{-1.42} & \textcolor{downred}{-2.40} & \textcolor{applegreen}{+3.77} & \textcolor{applegreen}{+7.59} & \textcolor{applegreen}{+7.98} & \textcolor{applegreen}{+4.40} & \textcolor{applegreen}{+1.47} & 0.00 \\
\hline
\multirow{7}{*}{\rotatebox{90}{b-1b7}} & Vanilla & 53.51 & 61.93 & \textbf{68.35} & \textbf{70.26} & 72.60 & \textbf{76.34} & 67.48 & 68.13 & 69.12 & \uline{70.83} \\
 & Keywords & 30.52 & 44.37 & 46.84 & 55.40 & 57.19 & 65.63 & 58.28 & 47.25 & 60.29 & 62.50 \\
 & Summary & 32.62 & 44.94 & 50.00 & 54.20 & 63.01 & 63.84 & 61.96 & \uline{72.53} & 70.59 & 62.50 \\
 & SelCon & 35.96 & 43.32 & 45.89 & 47.72 & 56.51 & 55.36 & 57.67 & 40.66 & 57.35 & 50.00 \\
 & LLMLingua & \uline{53.62} & \textbf{64.50} & \uline{67.72} & 68.59 & \uline{75.00} & \uline{75.45} & \uline{67.49} & 70.33 & \uline{75.00} & \uline{70.83} \\
 & \cellcolor{markcolor} CoCR-RAG & \cellcolor{markcolor}\textbf{55.78} & \cellcolor{markcolor}\uline{63.55} & \cellcolor{markcolor}\uline{67.72} & \cellcolor{markcolor}\textbf{70.26} & \cellcolor{markcolor}\textbf{80.48} & \cellcolor{markcolor}\uline{75.45} & \cellcolor{markcolor}\textbf{77.91} & \cellcolor{markcolor}\textbf{80.22} & \cellcolor{markcolor}\textbf{77.94} & \cellcolor{markcolor}\textbf{83.33} \\
 & $\Delta$ & \textcolor{applegreen}{+2.27} & \textcolor{applegreen}{+1.62} & \textcolor{downred}{-0.63} & 0.00 & \textcolor{applegreen}{+7.88} & \textcolor{downred}{-0.89} & \textcolor{applegreen}{+10.43} & \textcolor{applegreen}{+12.09} & \textcolor{applegreen}{+8.82} & \textcolor{applegreen}{+12.50} \\
\hline
\multirow{7}{*}{\rotatebox{90}{b-3}} & Vanilla & \uline{53.06} & 62.40 & \uline{69.30} & 67.63 & \uline{75.34} & \uline{74.55} & \uline{74.23} & \uline{73.63} & 79.41 & 83.33 \\
 & Keywords & 32.16 & 41.60 & 47.15 & 53.48 & 56.51 & 61.61 & 58.90 & 50.55 & 52.94 & 54.17 \\
 & Summary & 31.26 & 38.74 & 47.31 & 47.72 & 52.05 & 57.59 & 51.53 & 50.55 & 60.29 & 62.50 \\
 & SelCon & 44.11 & 51.53 & 54.75 & 59.71 & 65.75 & 68.30 & 66.87 & 64.84 & 72.06 & 66.67 \\
 & LLMLingua & 51.13 & \uline{62.69} & 67.25 & \uline{68.11} & 71.92 & 73.21 & 71.78 & 72.53 & \textbf{83.82} & \textbf{87.50} \\
 & \cellcolor{markcolor} CoCR-RAG & \cellcolor{markcolor}\textbf{58.38} & \cellcolor{markcolor}\textbf{66.13} & \cellcolor{markcolor}\textbf{70.25} & \cellcolor{markcolor}\textbf{72.90} & \cellcolor{markcolor}\textbf{78.42} & \cellcolor{markcolor}\textbf{77.68} & \cellcolor{markcolor}\textbf{79.75} & \cellcolor{markcolor}\textbf{75.82} & \cellcolor{markcolor}\textbf{83.82} & \cellcolor{markcolor}\textbf{87.50} \\
 & $\Delta$ & \textcolor{applegreen}{+5.32} & \textcolor{applegreen}{+3.73} & \textcolor{applegreen}{+0.95} & \textcolor{applegreen}{+5.27} & \textcolor{applegreen}{+3.08} & \textcolor{applegreen}{+3.13} & \textcolor{applegreen}{+5.52} & \textcolor{applegreen}{+2.19} & \textcolor{applegreen}{+4.41} & \textcolor{applegreen}{+4.17} \\
\hline
\multirow{7}{*}{\rotatebox{90}{L-7}} & Vanilla & 49.83 & 50.29 & 59.02 & 69.06 & 73.63 & 66.07 & 73.01 & 76.92 & \uline{85.29} & 75.00 \\
 & Keywords & 41.33 & 56.58 & 59.33 & 64.51 & 67.81 & 70.98 & 71.17 & 68.13 & 82.35 & 70.83 \\
 & Summary & 37.60 & 41.41 & 44.46 & 42.45 & 50.68 & 47.32 & 51.53 & 61.54 & 64.71 & 58.33 \\
 & SelCon & 53.06 & 61.26 & 66.14 & 67.39 & 73.63 & 74.11 & \textbf{77.91} & \uline{81.32} & 82.35 & \textbf{79.17} \\
 & LLMLingua & \textbf{61.16} & \uline{62.98} & \textbf{74.94} & \uline{76.02} & \uline{73.97} & \uline{76.79} & 70.55 & 80.22 & \textcolor{bestcolor}{\textbf{88.24}} & 75.00 \\
 & \cellcolor{markcolor} CoCR-RAG & \cellcolor{markcolor}\uline{60.82} & \cellcolor{markcolor}\textbf{67.37} & \cellcolor{markcolor}\uline{74.05} & \cellcolor{markcolor}\textcolor{bestcolor}{\textbf{78.66}} & \cellcolor{markcolor}\textbf{78.77} & \cellcolor{markcolor}\textbf{78.12} & \cellcolor{markcolor}\uline{76.69} & \cellcolor{markcolor}\textbf{85.71} & \cellcolor{markcolor}\uline{85.29} & \cellcolor{markcolor}\textbf{79.17} \\
 & $\Delta$ & \textcolor{applegreen}{+10.99} & \textcolor{applegreen}{+17.08} & \textcolor{applegreen}{+15.03} & \textcolor{applegreen}{+9.60} & \textcolor{applegreen}{+5.14} & \textcolor{applegreen}{+12.05} & \textcolor{applegreen}{+3.68} & \textcolor{applegreen}{+8.79} & 0.00 & \textcolor{applegreen}{+4.17} \\
\hline
\multirow{7}{*}{\rotatebox{90}{L-13}} & Vanilla & 54.64 & \uline{69.75} & 71.68 & 73.38 & 75.34 & 66.96 & 68.71 & 70.33 & 80.88 & \textbf{83.33} \\
 & Keywords & 43.15 & 53.53 & 58.54 & 62.59 & 68.49 & 72.77 & 74.23 & 65.93 & 80.88 & \textbf{83.33} \\
 & Summary & 38.73 & 46.09 & 44.78 & 40.04 & 46.58 & 47.77 & 53.37 & 62.64 & 60.29 & 45.83 \\
 & SelCon & 56.29 & 64.79 & 68.51 & 70.74 & 75.68 & 72.32 & 78.53 & \uline{82.42} & 82.35 & 79.17 \\
 & LLMLingua & \textcolor{bestcolor}{\textbf{61.83}} & 67.94 & \uline{75.00} & \uline{75.78} & \uline{79.79} & \uline{75.45} & \uline{79.75} & 76.92 & \textbf{85.29} & 79.17 \\
 & \cellcolor{markcolor} CoCR-RAG & \cellcolor{markcolor}\uline{60.25} & \cellcolor{markcolor}\textcolor{bestcolor}{\textbf{70.90}} & \cellcolor{markcolor}\textcolor{bestcolor}{\textbf{75.16}} & \cellcolor{markcolor}\textbf{77.46} & \cellcolor{markcolor}\textbf{80.82} & \cellcolor{markcolor}\textbf{79.46} & \cellcolor{markcolor}\textbf{80.98} & \cellcolor{markcolor}\textcolor{bestcolor}{\textbf{86.81}} & \cellcolor{markcolor}\uline{83.82} & \cellcolor{markcolor}\textbf{83.33} \\
 & $\Delta$ & \textcolor{applegreen}{+5.61} & \textcolor{applegreen}{+1.15} & \textcolor{applegreen}{+3.48} & \textcolor{applegreen}{+4.08} & \textcolor{applegreen}{+5.48} & \textcolor{applegreen}{+12.50} & \textcolor{applegreen}{+12.27} & \textcolor{applegreen}{+16.48} & \textcolor{applegreen}{+2.94} & 0.00 \\
\hline

\end{tabular}
\end{adjustbox}
\end{table}

\end{document}